\documentclass{nle}

\title[Lexical Resources To Improve Low-Resource Part-of-Speech Tagging]
      {The Best of Both Worlds: Lexical Resources To Improve Low-Resource Part-of-Speech Tagging}
\author[Plank, Klerke, Agi{\' c}]
       {Barbara Plank \quad Sigrid Klerke \quad {\v Z}eljko Agi{\' c}\\
	Department of Computer Science\\
    IT University of Copenhagen\\
    Rued Langgaards Vej 7, 2300 Copenhagen S, Denmark\\
    {\tt \{bplank,sikl,zeag\}@itu.dk}}

\received{November 15, 2018}

\pagerange{\pageref{firstpage}--\pageref{lastpage}}
\pubyear{2018}

\usepackage{graphicx}

\usepackage{booktabs}
\usepackage{url}
\usepackage{subfig}
\usepackage[export]{adjustbox}
\usepackage{todonotes}
\usepackage{alltt}
\usepackage{enumitem}

\usepackage{xcolor}

\begin{document}

\label{firstpage}
\maketitle

\begin{abstract}
In natural language processing, the deep learning revolution has shifted the focus  from conventional hand-crafted symbolic representations to dense inputs, which are adequate representations learned automatically from corpora. However, particularly when working with low-resource languages, small amounts of symbolic lexical resources such as user-generated lexicons are often available even when gold-standard corpora are not. Such additional linguistic information is though often neglected, and recent neural approaches to cross-lingual tagging typically rely only on word and subword embeddings. While these representations are effective, our recent work has shown clear benefits of combining the best of both worlds:  integrating conventional lexical information improves neural cross-lingual part-of-speech (PoS) tagging~\cite{dsds}. However, little is known on how complementary such additional information is, and to what extent improvements depend on the coverage and quality of these external resources. This paper seeks to fill this gap by providing the first thorough analysis on the contributions of lexical resources for cross-lingual PoS tagging in neural times.
\end{abstract}

\section{Introduction}

From the viewpoint of modern language technology, languages may be seen as ``dialects with a part-of-speech tagger, a treebank, and a machine translation system''~\cite{tiedemann2016synthetic}. 

Such fortunate dialects are a vast minority: At this point in time, only around 100 or 1\% of the world's spoken languages come bundled with a tagger. The remainder are severely low-resource, even if the need to process information that they convey is just as urgent. The sharp disconnect between the technological needs of the many and the resource wealth of the few has ushered in a vibrant research avenue in transfer learning for language technology. 

In cross-lingual learning, we strive to adapt basic models such as part-of-speech taggers to transfer them over from well-resourced source languages to deprived target languages. Since the best taggers are based on supervised learning from hand-annotated corpora, the cross-lingual learning efforts predominantly involve a degree of creativity to compensate for the lack of such corpora when enabling a language.

\subsection{Background}

Nowadays there are several groups of approaches to yield a workable part-of-speech tagger for a true low-resource language. The approaches all share a common theme: In absence of direct supervision through annotated corpora, they resort to replacement data that offers some lossy approximation of the desired but absent resource. Consequently, the ways to go about creating taggers ``almost from scratch'' mainly differ depending on what kind of approximation is readily available:
\begin{itemize}[noitemsep]
\item[--] {\bf Annotation projection: } If parallel text is available between source and target languages, one can follow~\cite{yarowsky2001inducing} and project annotations from sources into targets. It is fortunate that we can find reasonable sentence and word alignments via unsupervised learning. Thus, transferring taggers amounts to pushing the source tags through the alignments onto the target words, and then learning a target-language tagger on the outputs. While annotation projection may take many shapes, the core idea of exchanging labels through parallel corpora has been instantiated with great success across tasks, from creating word representations~\cite{ruder2017survey}, sequence labeling tasks such as part-of-speech tagging~\cite{das-petrov:2011,agic:ea:2015} and named entity recognition~\cite{ni2017weakly}, to syntactic and semantic parsing~\cite{rasooli2016cross,evang2016cross}. The received wisdom in the field dictates that for most low-resource language processing tasks, annotation projection is a reasonable first choice, albeit with some exceptions~\cite{mayhew2017cheap,enghoff2018low}.
\item[--] {\bf Type-level supervision: }In absence of word-level annotations, lexical resources such as dictionaries may offer a more distant supervision: Essentially, one could ``tag'' any running text using a dictionary of words and their parts of speech, and then come up with creative ways to make a model learn from such data. The learning signal is here limited: i) a given dictionary is very unlikely to offer full coverage of any running text, and ii) the dictionary entries are inherently ambiguous as, e.g., ``play'' can be both {\sc noun} and {\sc verb}. Learning under such incomplete supervision offers challenges that past work has attempted to surmount through semi-supervised techniques~\cite{das-petrov:2011,li-et-al:2012}, among others. There, the learner choice and quality of dictionaries is of vital importance, by contrast to annotation projection where emphasis is placed on yielding the target-language training data, which can then be fed to any fully supervised learner.
\item[--] {\bf Rapid annotation: }While large quantities of annotated data are costly to produce, smaller quantities may not be as prohibitive, especially if procured through a more noisy process. This idea is essential for a set of resource-lean annotation efforts, whereby small seeds of lower-quality training data are provisioned typically by crowdsourcing annotations to online workers~\cite{garrette-baldridge:2013:NAACL-HLT}. The challenge is to select an appropriate annotation task, such as corpus annotation or dictionary creation, and cast the task into workable units to ensure sufficient quality under annotator noise.
\item[--] {\bf Unsupervised learning}: Left as the last resort as such taggers typically do not yield workable results across diverse low-resource languages. On the other hand, recent work based on multilingual word representations or subword features does hold promise~\cite{zeman2016bible,stratos2016unsupervised}. Still, if the goal is to create a usable low-resource tagger, it is advisable to exhaust the previously listed methods before resorting to unsupervised learning. Either way, the focus of our contribution does not extend to unsupervised part-of-speech taggers.
\end{itemize}
In cross-lingual learning research, the above groups of methods are contrasted to one another. That is to say, for example, that a novel addition to annotation projection from multiple sources will typically be compared to all the other usual suspects: an unsupervised baseline, a fully supervised upper bound, and then prominent distant supervision approaches from the ``competing camp''.

However, if our goal is ultimately to create the best possible cross-lingual part-of-speech tagger for a severely under-resourced language, it may be worthwhile to violate such constraints of methodological purity, and to opt for a more pragmatic approach instead. Our proposal is one such ``violation'' for the benefit of low-resource languages.

\subsection{``Take what you can get'' in low-resource tagging}

Can we learn low-resource part-of-speech taggers by combining a multitude of various noisy information sources, e.g., projections and user-generated lexical resources, pre-trained word representations, character embeddings, and the like?

As it turns out, the research into this practical question is rather scarce. Very recently, a contribution by~\cite{dsds} explored how to integrate various disparate sources of distant supervision into a single robust neural tagger. We highlight their main findings:
\begin{itemize}[noitemsep]
\item[--] It is possible to seamlessly integrate various heterogeneous sources of distant supervision into an bidirectional recurrent neural network labeler~\cite{plank:ea:2016} for improved low-resource part-of-speech tagging.
\item[--] The resulting tagger {\sc DsDs} offers state-of-the-art scores in cross-lingual part-of-speech tagging across multiple languages.
\end{itemize}
The tagger by~\cite{dsds} offers a very simple but novel and most of all markedly beneficial way of integrating lexical resources into a neural sequence labeler. In a nutshell, it offers ``the best of both worlds'': a neural part-of-speech tagger in which the introduction of external lexical knowledge in a ``soft'' manner (via embedded linguistic knowledge) substantially improves tagging performance in low-resource languages, and is more robust than hard-decoding based type-level supervision approaches. Most notably, their contribution seems to suggest that the introduction of lexical resources enables the tagger to generalize even beyond simply extending the lexical coverage: Adding the noisy lexicon makes {\sc DsDs} perform significantly better, even on entirely unseen words, across many languages.

The research by~\cite{dsds} is a starting point for our contributions. What makes their neural tagger tick in such a way that it benefits from the inclusion of external lexical knowledge, even when it is user-generated and noisy?

\subsection{Our contribution: A lexically enriched neural tagger explained}

In this paper, we seek to understand under which conditions a low-resource neural tagger benefits from external lexical knowledge. Our main contribution is an in-depth analysis of a cross-lingual neural Part-of-Speech tagger enriched with symbolic lexical information. In particular: 

\begin{itemize}
\item[a)] we evaluate the neural tagger across a total of 25 languages comparing it to strong baselines, which either explicitly or implicitly use lexicon information (weakly supervised learning), extending the comparison of recent work~\cite{dsds} with an alternative baseline which uses retrofitting; 
\item[b)] we investigate the reliance of dictionary size and propose to sample new dictionaries by frequency; 
\item[c)] we analyze tagging performance across a range of factors, including tag set agreement, out-of-vocabulary items (with respect to lexicons or training data), ambiguity rate and frequency, and show results at an aggregate level, but more importantly also on the per-language level; finally,
\item[d)] we dissect model-internal representations; as the tagger learns to rely not only explicitly on the external resource, we investigate to what extent model-internal representations capture morphosyntactic information by investigating the learned encodings quantitatively via a similarity analysis as well as via introspective diagnostic classifiers (probing tasks).
\end{itemize}

We find a clear synergetic effect between a neural tagger and symbolic linguistic knowledge. The composition of the dictionary plays a more important role than its coverage, and enriching the tagger with the lexicons helps to generalize beyond its coverage. This  opens the path for better low-resource taggers.

\section{Methodology}

This section first outlines the base neural tagging architecture.  It then depicts its extension to integrate lexical information, followed by details on the annotation projection step where we provide details on a novel and effective coverage-based data selection method. The section ends with details on the experimental setup.

\subsection{Tagging model} 

Sub-word and especially character-level modeling is currently pervasive in top-performing neural sequence taggers, owing to its capacity to effectively capture morphological features that are useful in labeling out-of-vocabulary (OOV) items. As exemplified by~\cite{plank:ea:2016}, sub-word information is often coupled with standard word embeddings to mitigate OOV issues. Specifically, i) word embeddings are typically built from massive unlabeled datasets and thus OOVs are less likely to be encountered at test time, while ii) character embeddings offer further linguistically plausible fallback for the remaining OOVs through modeling intra-word relations. Through these approaches, multilingual PoS tagging has seen tangible gains from neural methods in the recent years. We use a bidirectional long short-term memory network (bi-LSTM)~\cite{graves:schmidhuber:2005,Hochreiter:Schmidhuber:97,plank:ea:2016} with a rich word encoding model which consists of a recurrent subword-based representation paired with pre-trained word embeddings. 

\begin{figure}\center
\includegraphics[width=0.45\textwidth]{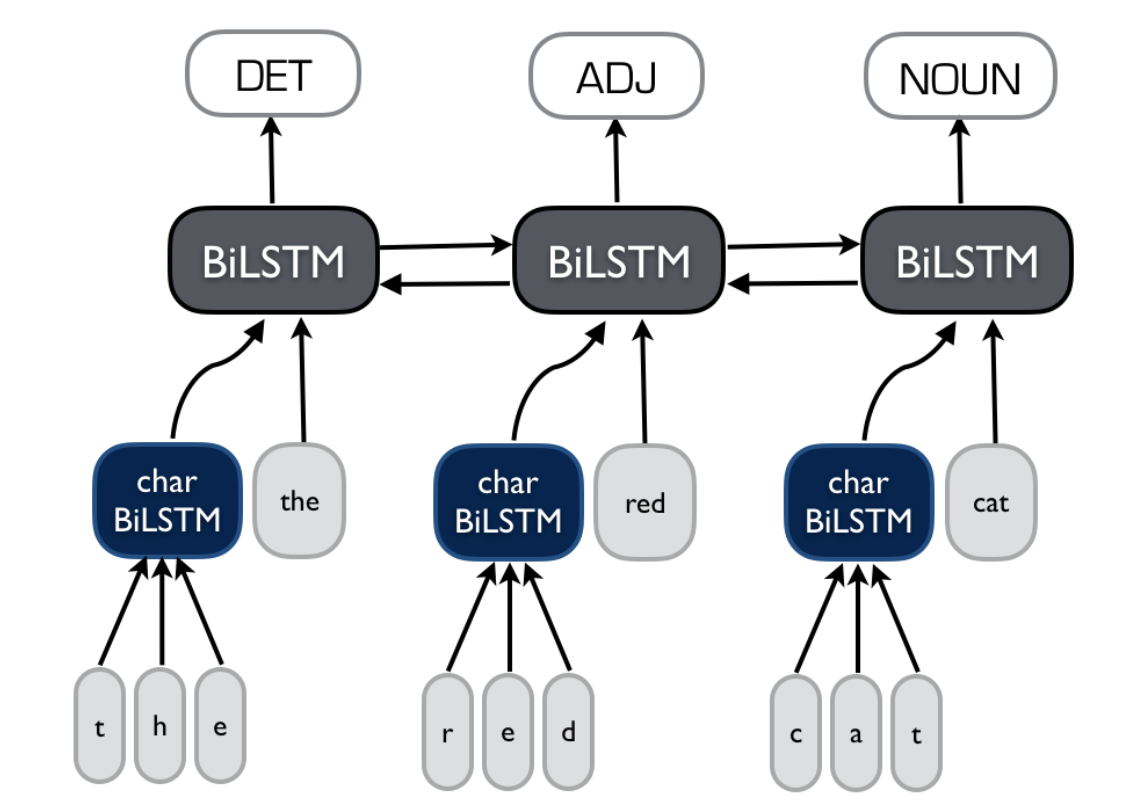}
\includegraphics[width=0.5\textwidth]{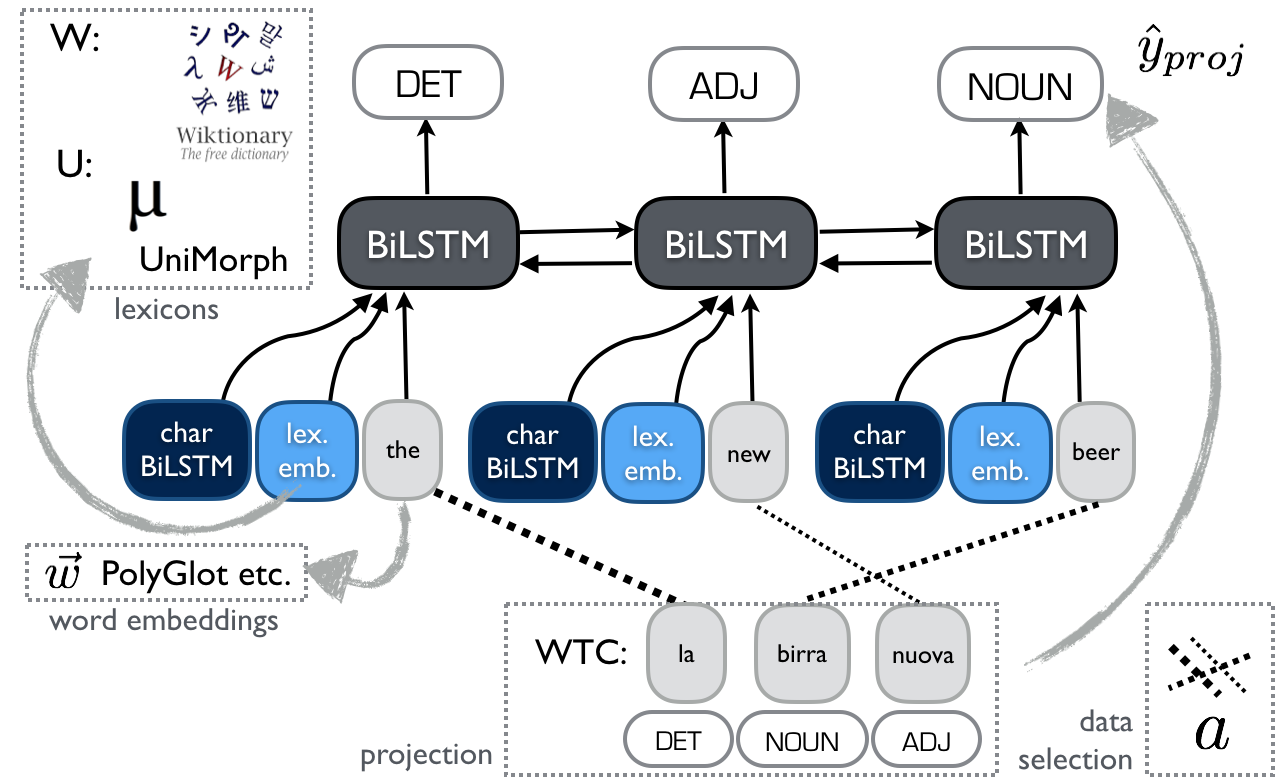}
\caption{a) \textbf{Left}: Illustration of the base tagging model. b) \textbf{Right:} \textsc{DsDs} (Distant Supervision from Disparate Sources) tagger, figure from Plank \& Agi{\' c} (2018).}
\label{fig:model}
\end{figure}

In more detail, following the notation by~\cite{kiperwasser2016}, let $x_{1:n}$ be a sequence of input vectors. In the base model, illustrated in Figure~\ref{fig:model} left, the input sequence $x_{1:n}$ consists of input word embeddings $\vec{w}$ and the two output (forward and backward, $\vec{c}_f$  and $\vec{c}_b$) states of a character-level bi-LSTM at the lower level, i.e.,  $\vec{cw} = \vec{c}_f \circ \vec{c}_b$, the concatenation of the character bi-LSTM end states, as originated in~\cite{plank:ea:2016}. Given $x_{1:n}$ and a
desired index {$i$}, the function $BiRNN_\theta(x_{1:n}, i)$ (in our case instantiated as LSTM) reads the input sequence in forward and reverse order, respectively, and uses the concatenated ($\circ$) output states of $\vec{w}$ and $\vec{cw}$ as input for tag prediction at position $i$. The prediction of the PoS tag at position $i$ is based on local argmax decoding. Global CRF decoding did not consistently improve PoS accuracy, as recently also independently noted by~\cite{yang:2018:coling}.

The tagger which we analyze in this paper is an extension of the base tagger, called \textit{distant supervision from disparate sources} (\textsc{DsDs}) tagger~\cite{dsds}. It is trained entirely on projected data (see Section~\ref{sec:projection}). \textsc{DsDs} is illustrated in Figure~\ref{fig:model} (right), and further differs from the base tagger by the integration of lexicon information into the model, as described next (Section~\ref{sec:resources}).

\subsection{Lexical resources}\label{sec:resources}

Dictionaries are a useful source for distant supervision~\cite{li-et-al:2012,tackstrom:ea:2013}. There are several ways to exploit such information: 
\begin{itemize}
\item[i)] as type constraints during encoding~\cite{tackstrom:ea:2013}, 
\item[ii)] to guide unsupervised learning~\cite{li-et-al:2012}, or 
\item[iii)] as additional signal during training.  
\end{itemize} 
\textsc{DsDs} takes the latter approach, and integrates lexical knowledge into the neural model by \textit{embedding} lexical features. That is, a lexicon embeddings vector is created and concatenated to the word and character-level representations at position $i$ (blue box in Figure~\ref{fig:model}): $\vec{w} \circ \vec{cw} \circ \vec{e}$.

In particular, given a lexicon $src$, we use  $\vec{e}_{src}$ to embed the lexicon into an $l$-dimensional space, where $\vec{e}_{src}$ is the concatenation of all embedded $m$ properties of length $l$ (empirically set, see Section~\ref{sec:setup}), and a zero vector for words not in the lexicon.  We found this simple approach to perform best by tuning on the dev set, and concatenation outperformed mean vector representations. 
Moreover,  \textit{embedding} the lexicon rather than using it as $n$-hot encoding as suggested earlier~\cite{sagot-martinezalonso:2017:IWPT} turns out to work better in practice~\cite{dsds}.

We compare \textsc{DsDs} to  alternative ways of using lexical information. The first approach uses lexical information directly at tagging time, the second approach is more implicit and uses the lexicon to induce better word embeddings for tagger initialization. In particular, besides comparing to previous work, we compare our model to two baselines:  a) a type-constraint approach during decoding~\cite{tackstrom:ea:2013}; and b) using the dictionary for retrofitting off-the-shelf embeddings~\cite{retrofitting} and initializing the tagger with those. The latter is a novel approach which, to the best of our knowledge, has not yet been evaluated in the neural tagging literature. The idea is to bring the off-the-shelf embeddings closer to the PoS tagging task by retrofitting the embeddings with syntactic clusters derived from the lexicon.

\subsubsection{Lexicons} We use linguistic resources that are user-generated and available for many languages. In this work, we focus on two such linguistic resources. The first  is \textsc{Wiktionary}, a word type dictionary that maps words to one of the 12 Universal PoS tags~\cite{li-et-al:2012,petrov:ea:2012}. Our second resource is \textsc{UniMorph}, a morphological dictionary that provides  inflectional paradigms across 350 languages~\cite{KIROV16.1077}. For Wiktionary, we use the freely available dictionaries from~\cite{li-et-al:2012} and \cite{agic:ea:2017}. UniMorph covers between 8-38 morphological properties (for English and Finnish, respectively).\footnote{More details: \url{http://unimorph.org/}}
Examples from these two resources are given in Figure~\ref{fig:examples}.

As we take a pragmatic ``all we can get approach'', we first experiment with using the dictionaries as they come. This implies that the sizes of the dictionaries vary considerably,  from a few thousand entries (e.g., for Hindi and Bulgarian) to 2M entries (Finnish UniMorph). The sizes  for the two linguistic resources are provided in the first two columns of Table~\ref{tbl:results}. As having a huge dictionary at disposal is not always realistic, we study  the impact of smaller dictionary sizes in Section~\ref{sec:sampling}.

\begin{figure}
\begin{alltt}
studio          NOUN,VERB
studioso        ADJ
\end{alltt}
\begin{alltt}
allenare        V;NFIN
allenavo        V;IND;PST;1;SG;IPFV
allenate        V;IND;PRS;2;PL
\end{alltt}
\caption{Example entries from Wiktionary and Unimorph, respectively.}
\label{fig:examples}
\end{figure}

Moreover, we take a deeper look at the quality of the lexicons by comparing tag sets to the gold treebank data, inspired by~\cite{li-et-al:2012}. In particular, let $T$ be the dictionary derived from the gold treebank (development data), and $W$ be the user-generated dictionary, i.e., the respective Wiktionary (as we are looking at PoS tags). For each word type, we compare the tag sets in $T$ and $W$ and distinguish six cases:

\begin{enumerate}
\item \textsc{None}: The word type is in the training data but not in the lexicon (out-of-lexicon).
\item \textsc{Equal}: $W = T$
\item \textsc{Disjoint}: $W \cap T = \emptyset$
\item \textsc{Overlap}: $W \cap T \ne \emptyset$
\item \textsc{Subset}: $W \subset T $ 
\item \textsc{Superset}: $W \supset T $
\end{enumerate}

In an ideal setup, the dictionaries contain no disjoint tag sets, and larger amounts of equal tag sets or superset of the treebank data. This is particularly desirable for approaches that take lexical information as type-level supervision. 

To build taggers for novel languages, we resort to annotation projection, as described next.

\subsection{Annotation projection} 
\label{sec:projection}

To produce noisy text annotations for a low-resource target language, we rely on one of the most prevalent approaches to cross-lingual learning, annotation projection, whereby we project sequential labels from source to target languages~\cite{yarowsky2001inducing}. The only requirement imposed by projection is that parallel texts are readily available between the languages to obtain sentence and word alignments, and that the source side is annotated for PoS. Reliable sentence and word alignments fortunately yield to unsupervised approaches.

\begin{figure}
\includegraphics[width=\textwidth]{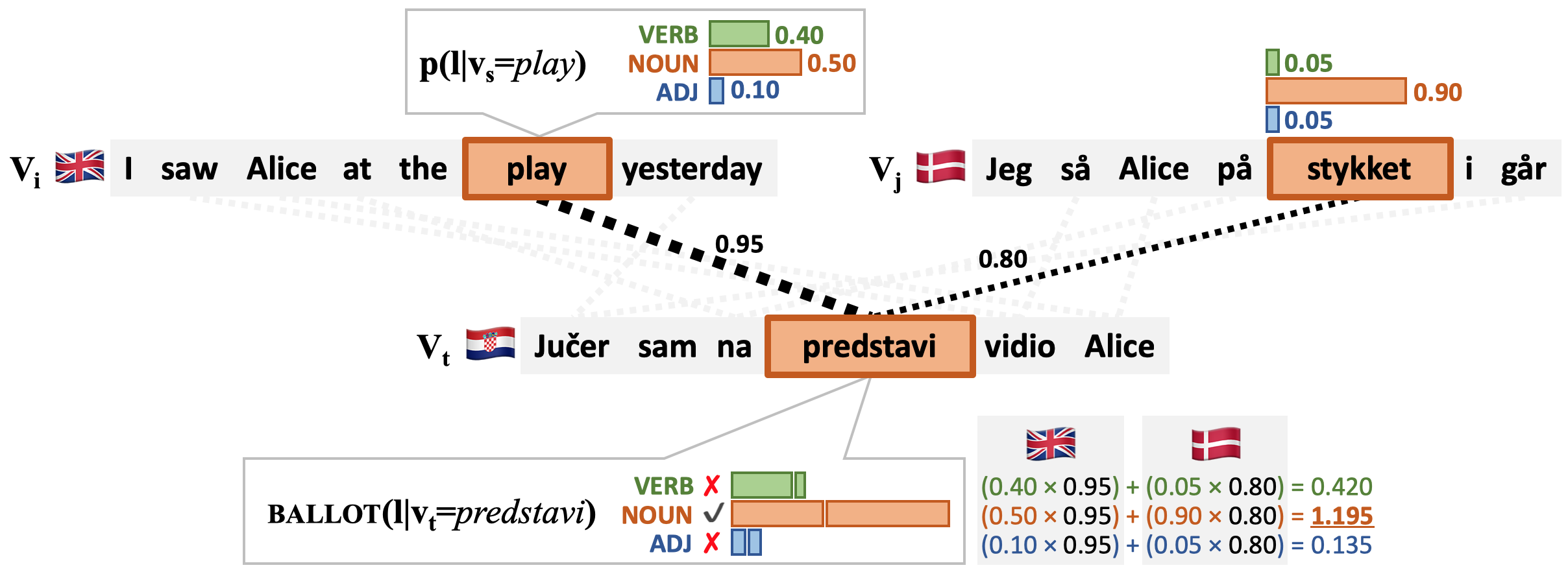}
\caption{An illustration of part-of-speech tag projection from two source sentences (English as $V_i$ and Danish as $V_j$) to one target (Croatian as $V_t$). In this example, the voting on tags is weighted by tagger confidences $p(l|v_s)$ and alignment probabilities $a(v_s, v_t)$. Most tags are omitted for simplicity, as we keep only three: {\sc adj, noun, verb}. The figure is adapted from Enghoff et al. (2018), who employ it for named entity label projection.}
\label{fig:projection}
\end{figure}

Here, we employ the approach by~\cite{agic2016multilingual}, where labels are projected from multiple sources to multiple targets and then decoded through weighted majority voting with word alignment probabilities and source PoS tagger confidences. The projection process is depicted in Figure~\ref{fig:projection}.

Specifically, our starting point is a parallel corpus, ideally large in cross-linguistic breadth. Such a corpus is basically a large collection of multilingual sentences. Each multilingual sentence is represented as a graph $G=(V, A)$ that contains the target sentence $t$ together with $n$ source sentences. The vertices $V=\cup_{j=0}^{n}V_j$ are the words: $v_t \in V_0, V_0 = V_t$ are the target words, while all other words $v_s \in V_i, 1 \leq i \leq n$ are the source words of the respective sources $i$. The edges $A$ in the multilingual sentence graph represent word alignments; their weights $a(v_s, v_t) \in (0, 1)$ are word aligner confidences. The graph is bipartite between source and target words $V_s, \forall s$ and $V_t$. Each source word carries a label distribution $p(l|v_s)$ that comes from a source-language PoS tagger and indicates the confidence of that tagger over PoS labels $l \in L$ for any given source word $v_s$. The label set $L$ in our experiment are the Universal PoS tags of~\cite{petrov:ea:2012}.

With this collection of multilingual sentences in place, for each aligned target word $v_t$ across the parallel corpus, we collect the tag contributions of the individual source words $v_s$ that are aligned with $v_t$ into a ballot:
$$
\textsc{ballot}(l|v_t) = \sum\limits_{v_s \in V_s}p(l|v_s)a(v_s, v_t).
$$
The votes are collected for each label $l$. Each source word $v_s$ casts a vote that amounts to its tagger's confidence $p(l|v_s)$ in that label, modulo the alignment confidence of that source word to the target word: $a(v_s, v_t)$. Once the ballot is full, i.e., when all aligned target words have received the votes from all the sources for all the labels, we can decode the labels:
$$
\textsc{label}(v_t) = \arg\max\limits_l \textsc{ballot}(l|v_t).
$$
Thus, for each target word $v_t$ we select the label $l$ that received the most votes across sources. The above formalization of the projection process is based on the work by~\cite{agic2016multilingual,enghoff2018low}, who employ it to project parts of speech and named entity labels, respectively.

We exploit the wide-coverage Watchtower corpus (WTC) by~\cite{agic2016multilingual}, in contrast to the typically used Europarl data. Europarl covers 21 languages of the EU with 400k-2M sentence pairs, while WTC spans 300+ widely diverse languages with only 10-100k pairs, in effect sacrificing depth for breadth, and introducing a more radical domain shift. However, as it turns out little projected data turns out to be the most beneficial~\cite{dsds}, reinforcing breadth for depth.

\begin{figure}
\includegraphics[width=0.5\textwidth]{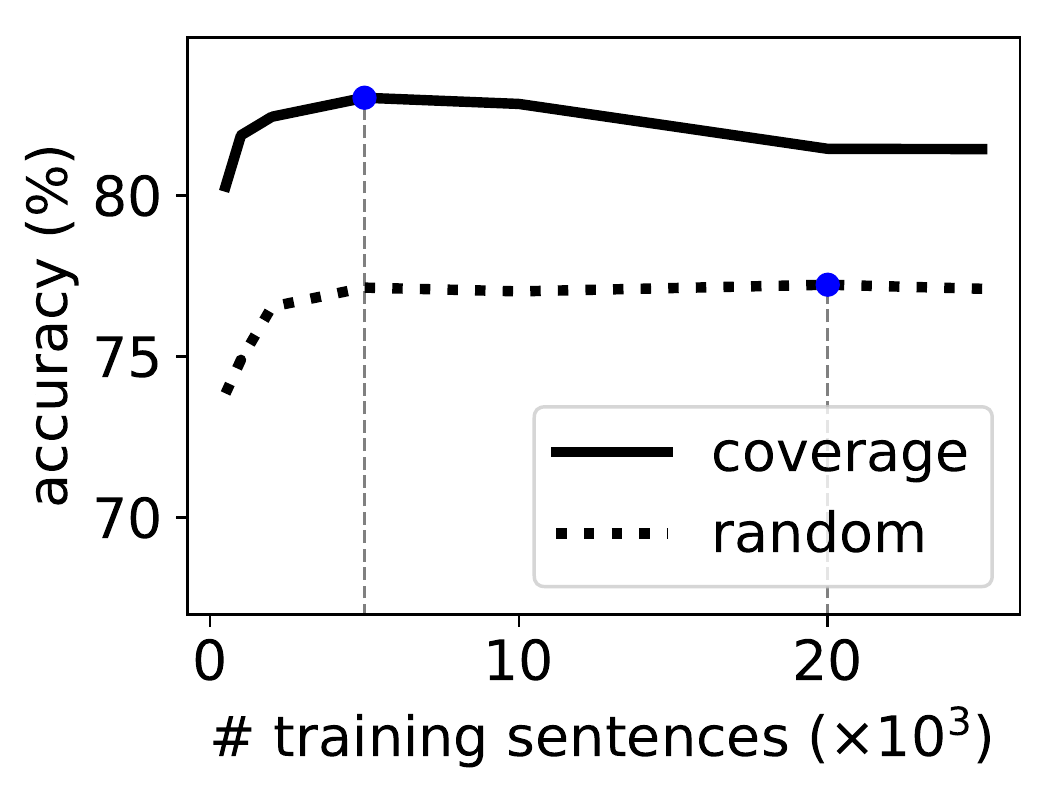}
\caption{Learning curve for random vs. coverage-based sentence selection in annotation
projection. Figure reported from Plank \& Agi{\' c} (2018). The benefits of selection are clear, as we reach a much higher tagging accuracy with a significantly smaller number of projected sentences.}
\label{fig:selection}
\end{figure}

Originally,~\cite{agic2016multilingual} employed 20k projected sentences at random to train their taggers. Here, we use a novel and more effective alternative: data selection by {\em alignment coverage}~\cite{dsds}. That is, target sentences are ranked by percentage of words covered by word alignment from 21 sources of~\cite{agic2016multilingual}, and we select the top $k$ covered instances for training. Since each target sentence $t$ may be aligned to any of the source sentences, we first express the coverage of $t$ by a single source sentence $i$:
$$
c_{i, t}=\frac{|\{v_t: \exists v_s, v_s \in V_i, a(v_s, v_t) \in A\}|}{|V_t|}, \forall i.
$$
The coverage $c_{i, t}$ is simply the percentage of words $v_t$ that have alignments coming in from source words of the source sentence $i$. Finally, we employ the mean coverage ranking of target sentences, whereby each target sentence is coupled with the arithmetic mean of the 21 individual word alignment coverages for each of the $n=21$ source-language sentences:
$$
c_t = \frac{1}{n}\sum\limits_{i=1}^{n}c_{i, t}.
$$
One alternative to mean coverage $c_t$ that we inspected to loosen the constraint was coverage by {\em any} source: $\hat{c}_t=|\{{v_t: \exists v_s, a(v_s, v_t) \in A}\}| / |V_t|$. In our development set experiments, however, the stricter constraint of mean coverage $c_t$ worked much better than a more relaxed any-source coverage.

Coverage-based instance selection yields substantially better cross-lingual taggers. As shown in Figure~\ref{fig:selection}, across all 21 languages, we learn better taggers with significantly fewer training instances (5k). Data selection outperforms random selection (+5\% absolute accuracy for the 5k setup). Training on 5k instances results in a sweet spot; more data (10k) starts to decrease performance, at a cost of runtime. In the rest of the paper,  we use the setup with 5k projected training instances as our base tagger, obtaining an average accuracy of 83.4\% (mean over 21 languages).

\subsection{Experimental setup}\label{sec:setup}
In this section we describe the baselines, the data and the tagger hyperparameters.

\subsubsection{Baselines} We compare to the following weakly-supervised PoS taggers:
\begin{itemize}
\item \textsc{\textbf{Agi{\' c}:}} Multi-source annotation projection with Bible parallel data by \cite{agic:ea:2015}.
\item \textsc{\textbf{Das:}} The label propagation approach by \cite{das-petrov:2011} over Europarl data.
\item \textsc{\textbf{Garrette:}} The approach by \cite{garrette-baldridge:2013:NAACL-HLT} that works with projections, dictionaries, and unlabeled target text.
\item \textsc{\textbf{Li:}} Wiktionary supervision~\cite{li-et-al:2012}.
\item \textsc{\textbf{T\"{a}ckstr\"{o}m:}} The type-token approach that works with projection over Europarl data~\cite{tackstrom:ea:2013}.

\end{itemize}

\subsubsection{Data}  In all experiments we work with the 12 Universal PoS tags~\cite{petrov:ea:2012}: ADJ, ADP, ADV, CONJ, DET, NOUN, NUM, PRON, PRT, VERB, X and .\ (punctuation). Our set of languages is motivated by accessibility to embeddings and dictionaries. For development, we use 21 dev sets of the Universal Dependencies 2.1 \cite{ud21}, and map them to the twelve PoS tags. We employ UD test sets on additional languages as well as the test sets of \cite{agic:ea:2015} to facilitate comparison to prior work. The test sets are a mixture of CoNLL \cite{buchholz-marsi:2006:CoNLL-X,nivre-EtAl:2007:EMNLP-CoNLL2007} and HamleDT test data \cite{zeman2014hamledt}, and are more distant from the training and development data. 
 
\subsubsection{Hyperparameters and off-the-shelf embeddings} We use the same hyperparameter setup as~\cite{dsds}, i.e., 10 epochs, word dropout rate ($p$=$.25$) and $l=40$-dimensional lexicon embeddings for \textsc{DsDs}, except for down-scaling the hidden dimensionality of the character representations from 100 to 32 dimensions. This ensures that our probing tasks always get the same input dimensionality: 64 (2x32) dimensions for $\vec{cw}$, which is the same dimension as the off-the-shelf embeddings.  Language-specific hyperparameter tuning could in theory lead to optimized models for each language. However, we use identical settings for each language which worked well and is less expensive, similar to~\cite{bohnet2018morphosyntactic}. For all experiments, we average over 3 randomly seeded runs, and provide mean accuracy.

We use the off-the-shelf Polyglot word embeddings~\cite{polyglot}, which performed consistently better than FastText~\cite{bojanowski2016enriching}. Word embedding initialization provides a consistent and considerable boost in this cross-lingual setup, up to 10\% absolute improvements across 21 languages when only 500 projected training instances are available~\cite{dsds}. Note that we empirically find it to be best to not update the word embeddings further in this noisy training setup, as that results in better performance, see Section~\ref{sec:updatingembeds}.

\section{Results}
\label{sec:results}

\subsection{Inclusion of lexical information} 

Table~\ref{tbl:results} presents tagging accuracy for the 21 individual languages on the development data, with means over all languages and language families (for which at least two languages are available). There are several take-aways.

\begin{table*}[ht!]
\centering
\caption{Results on the development sets. \textsc{Lex}: Size (word types) of dictionaries (W: Wiktionary, U: UniMorph). TC$_W$: type-constraints using Wiktionary; \textsc{DsDs}: our model. Best result in boldface. Averages over language families (with two or more languages in the sample, number of languages in parenthesis). } 
\resizebox{0.9\textwidth}{!}{
\begin{oldtabular}{lrr|rrrr}
\toprule
& \multicolumn{2}{c|}{\textsc{Lex} ($10^3$)} & \multicolumn{4}{c}{\textsc{Dev sets (UD2.1)}}             \\
\textsc{Language} & \textsc{W} & \textsc{U} & 5k & TC$_W$ & \textsc{Retro} & \textsc{DsDs} \\
\midrule
Bulgarian (bg) & 3 & 47 & 89.8 & 89.9 & 87.1 & \textbf{91.0} \\
Croatian (hr) & 20 & -- & 84.7 & 85.2 & 83.0 & \textbf{85.9} \\
Czech (cs) & 14 & 72 & \textbf{87.5} & \textbf{87.5}& 84.9 & 87.4 \\
Danish (da) & 22 & 24 &  89.8 & 89.3 & 88.2 & \textbf{90.1} \\
Dutch (nl) & 52 & 26 & 88.6 & 89.2 & 86.6 & \textbf{89.6} \\
English (en) & 358 & 91 & 86.4 & \textbf{87.6} & 82.5 & 87.3  \\
Finnish (fi) & 104 & 2,345 & 81.7 & 81.4 & 79.2 & \textbf{83.1} \\
French (fr) & 17 & 274 & \textbf{91.5} & 90.0 & 89.8 & 91.3 \\
German (de) & 62 & 71 & 85.8 & 87.1 & 84.7 & \textbf{87.5}\\
Greek (el) & 21 & -- & 80.9 & \textbf{86.1} &79.3& 79.2\\
Hebrew (he) & 3 & 12 & 75.8 & 75.9 & 71.7 & \textbf{76.8}\\
Hindi (hi) & 2 & 26 & 63.8 & 63.9 & 63.0 & \textbf{66.2}\\
Hungarian (hu) & 13 & 13 & \textbf{77.5} & \textbf{77.5} & 75.5 & 76.2 \\
Italian (it) & 478 & 410 & 92.2 & 91.8 & 90.0 & \textbf{93.7} \\
Norwegian (no) & 47 & 18 & 91.0 & 91.1 & 88.8 & \textbf{91.4}\\
Persian (fa) & 4 & 26 &  43.6 & 43.8 & \textbf{44.1} & 43.6\\ 
Polish (pl) & 6 & 132 & 84.9 & 84.9 & 83.3 & \textbf{85.4}\\
Portuguese & 41 & 211 & 92.4 & 92.2 & 88.6 & \textbf{93.1}\\
Romanian (ro) & 7 & 4 & 84.2 & 84.2 & 80.2 & \textbf{86.0} \\
Spanish (es) & 234 & 324 & 90.7 & 88.9 & 88.9 & \textbf{91.7} \\
Swedish (sv) & 89 & 67 & 89.4 & 89.2 & 87.0 & \textbf{89.8}\\
\midrule
 \textsc{Avg(21)} & & & 83.4 & 83.6 & 81.3 & \textbf{84.1}\\
\midrule
\textsc{Germanic (6)} & & & 88.5 & 88.9 & 86.3 & \textbf{89.3} \\
\textsc{Romance (5)} & & & 90.8 & 90.1 & 88.4 & \textbf{91.4}\\
\textsc{Slavic (4)} & & & 86.7 & 86.8 & 84.6 & \textbf{87.4}\\
\textsc{Indo-Iranian (2)} & & & 53.7 & 53.8 & 53.5 & \textbf{54.9} \\
\textsc{Uralic (2)} &  & & 79.6 & 79.4 & 79.2 &  \textbf{79.6}\\
\bottomrule
\end{oldtabular}
}
\label{tbl:results}
\end{table*}

Combining the best of two worlds results in the overall best tagging accuracy. Embedding lexical information into a neural tagger improves tagging accuracy from 83.4 to 84.1 (means over 21 languages). This replicates recent findings~\cite{dsds}, but with an altered character representation size setup.

On 15 out of 21 languages, \textsc{DsDs} is the best performing model. On two languages, type constraints work the best (English and Greek). The best performance is achieved only on one language (Persian) by retrofitting, which is a more implicit use of the lexicon, but this is the language with the overall lowest performance. On three languages, Czech, French and Hungarian, the baseline remains the best model, none of the lexicon-enriching approaches works. We proceed to inspect these results in more detail.

\begin{figure}[h!]
   \includegraphics[width=.85\textwidth]{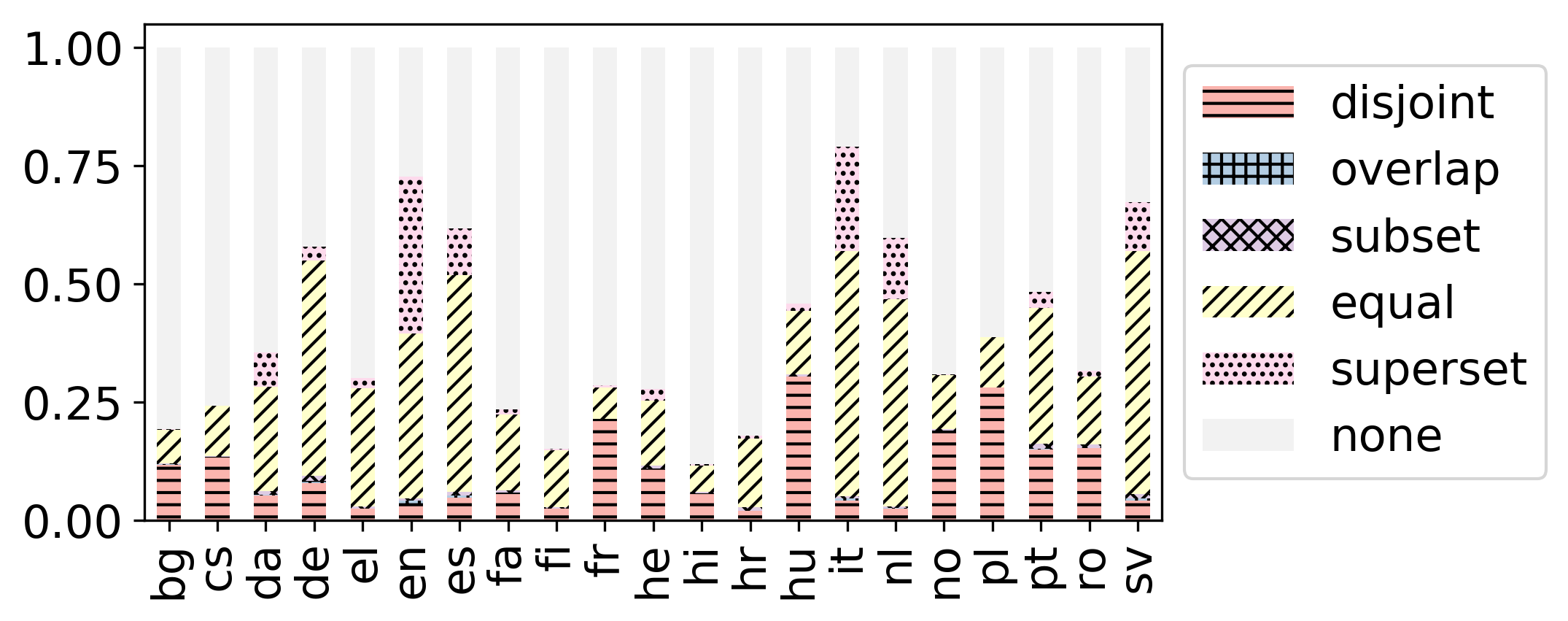}
   \caption{Analysis of Wiktionary vs gold (dev set) tag sets. `None': percentage of word types not covered in the lexicon. `Disjoint' means that the gold data and Wiktionary do not agree on the tag sets. See Section~\ref{sec:resources} for details on other categories. }
   \label{fig:tagset-typelevel-all}
 \end{figure}

Overall, type-constraints improve the baseline but only slightly (83.4 vs 83.6). Intuitively, this more direct use of lexical information requires the resource to be high coverage and a close fit to the evaluation data, to not introduce too many pruning errors during decoding due to contradictory tag sets. To analyze this, we look at the tag set agreement in Figure~\ref{fig:tagset-typelevel-all}. For languages for which the level of \textit{disjoint} tag set information is low, such as Greek, English, Croatian, Finnish and Dutch, type constraints are expected to help. This is in fact the case, but there are exceptions, such as Finnish. However, coverage of the lexicon is also important here, and for this morphologically rich language, the coverage is amongst the lowest (i.e., indicated by the large amount of the `none' category in Figure~\ref{fig:tagset-typelevel-all}). 

The more implicit use of lexical information in \textsc{DsDs} is more beneficial; it improves over the baseline and outperforms type constraints (83.6 vs 84.1). It helps on languages with relatively high dictionary coverage and low tag set disagreement, such as Danish, Dutch and Italian. Compared to type constraints, embedding the lexicon also helps on languages with low dictionary coverage, such as Bulgarian, Hindi, Croatian and Finnish, which is very encouraging and in sharp contrast to type constraints. The only outlier remains Greek, where it hurts. 

\begin{figure}[h!]
  \subfloat[a][Absolute improvement versus number of dictionary properties ($\rho$=0.08).]{
    \includegraphics[width=.65\textwidth]{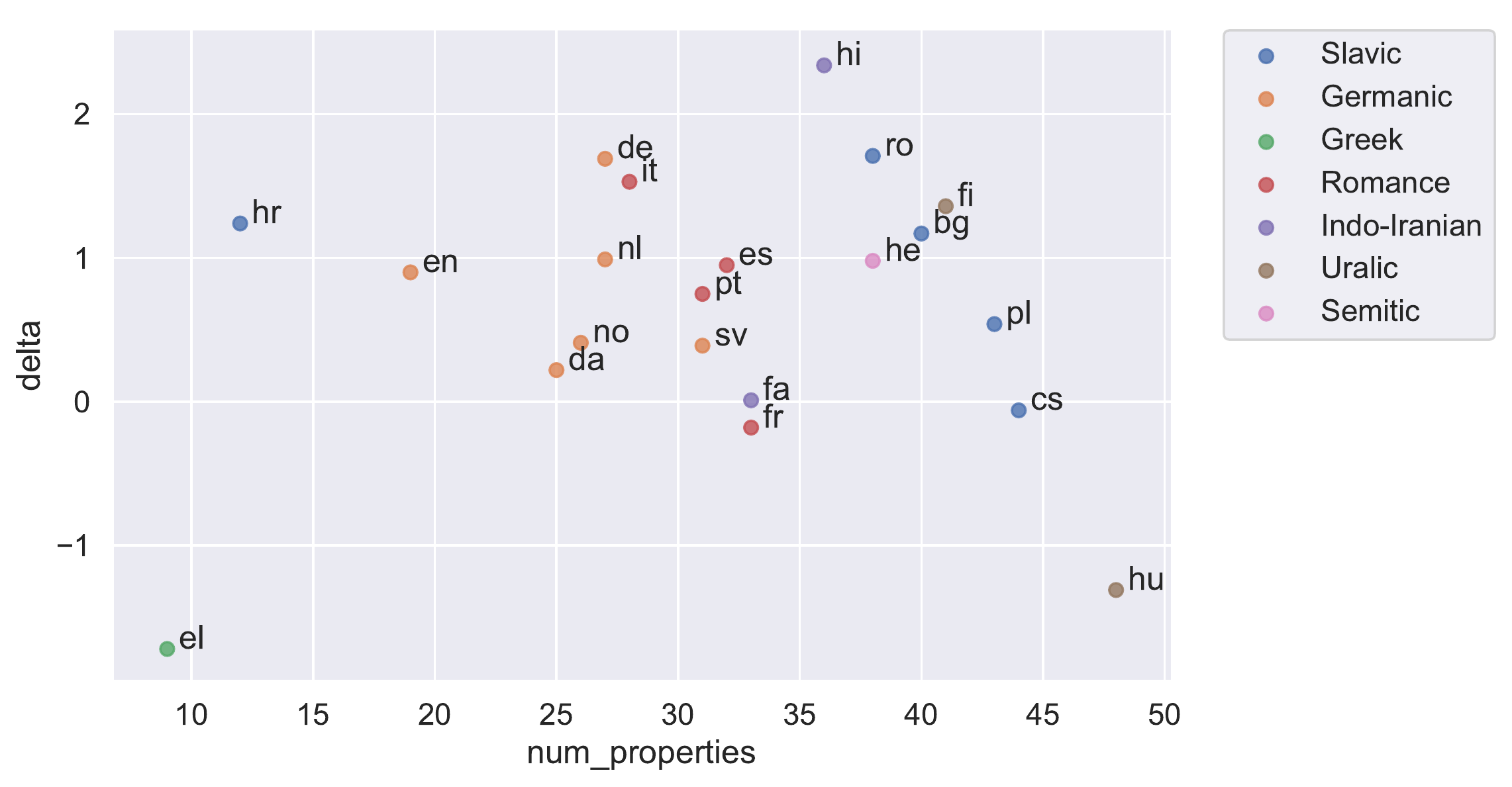}}
     \subfloat[b][OOVs (21 languages).]{            
    \includegraphics[width=.35\textwidth]{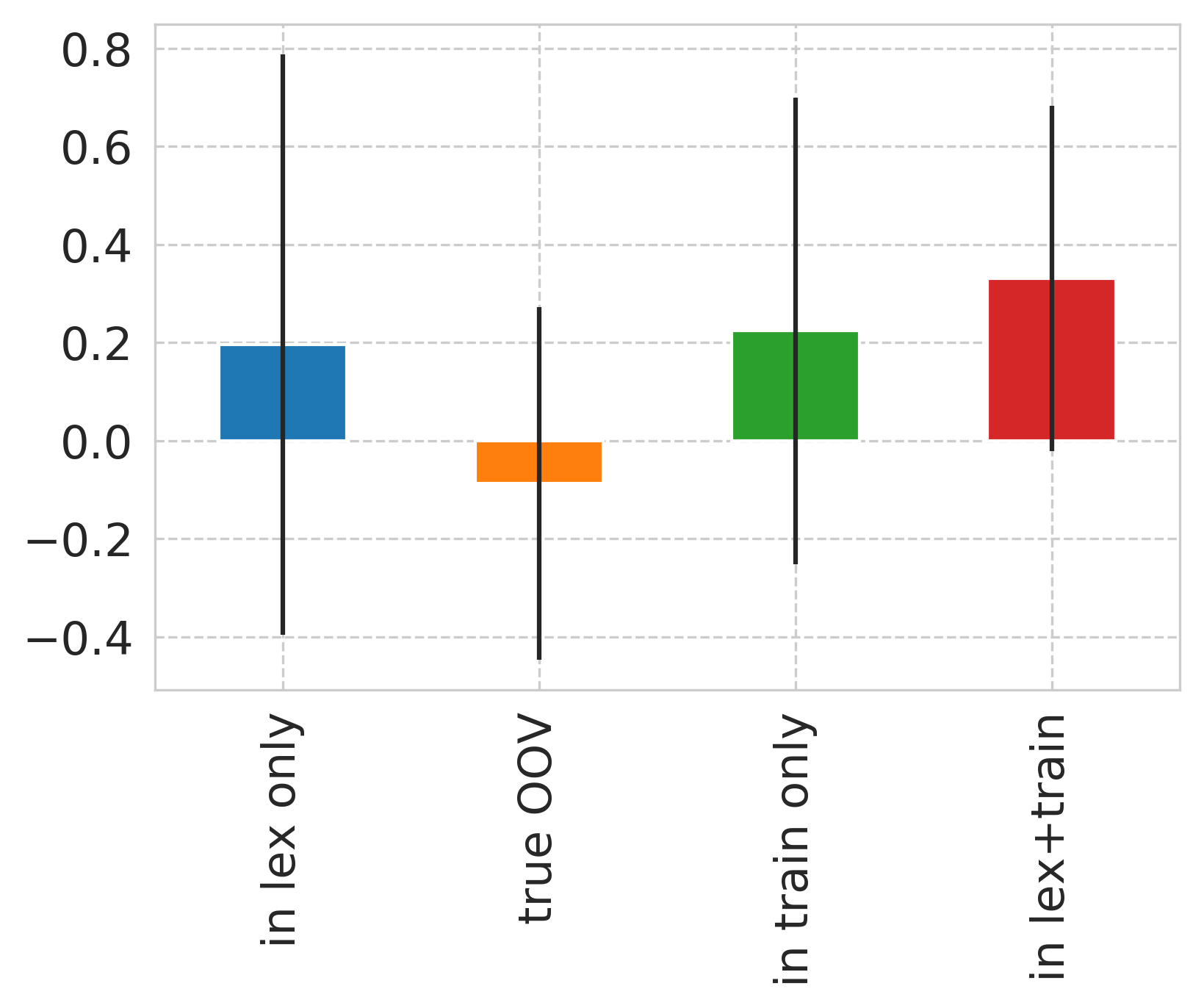}}
      \qquad
  \subfloat[c][Per language analysis: absolute improvements of \textsc{DsDs} over the baseline for words in the lexicon, in the training data, in both or in neither (true OOVs). ]{
    \includegraphics[width=\textwidth]{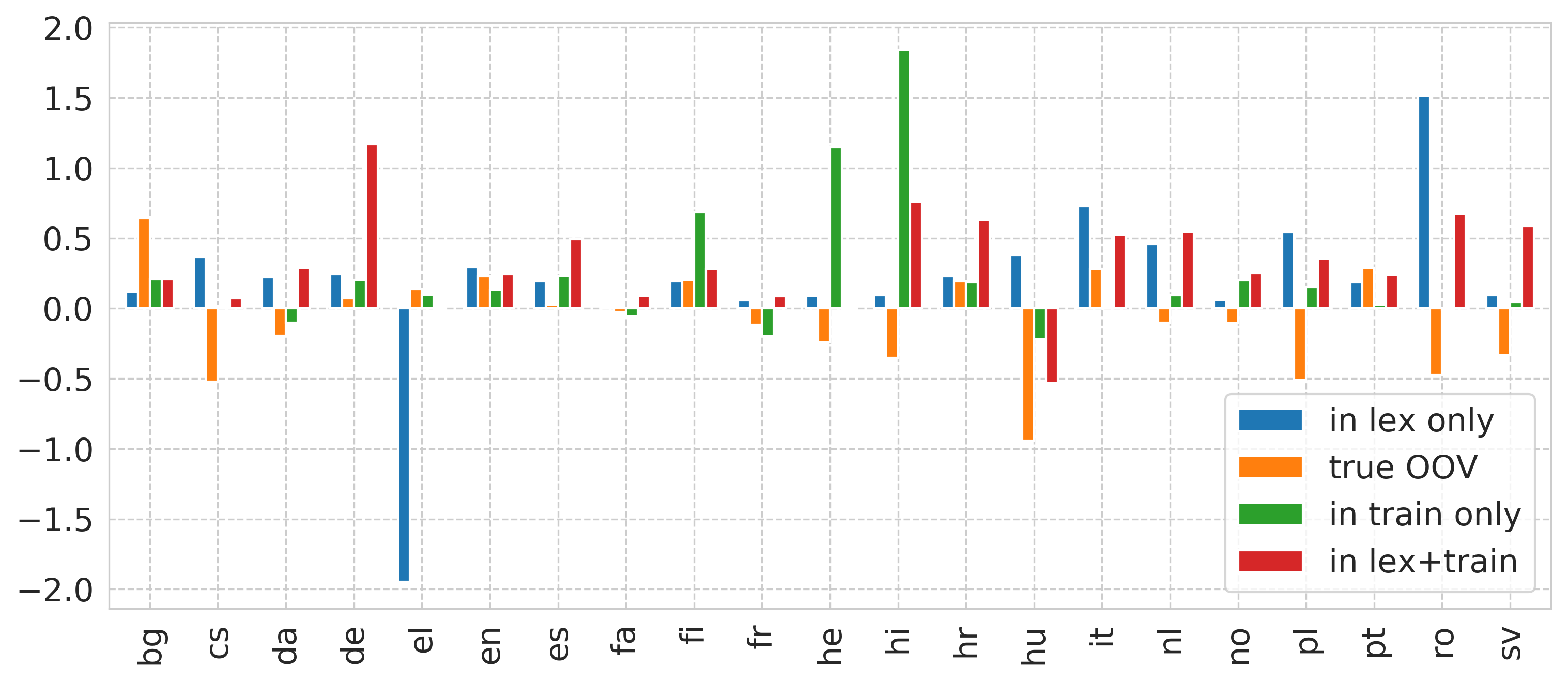}}
   \caption{Analysis of OOVs and dictionary properties.}
   \label{fig:oov-properties}
 \end{figure}

Figure~\ref{fig:oov-properties} plots the absolute improvement in tagging accuracy over the baseline versus the number of properties in the dictionaries.  Slavic and Germanic languages cluster nicely, with some outliers (Croatian, Hungarian). However, there is only a weak positive correlation ($\rho$=0.08). More properties do not necessarily improve performance, and lead to sparsity. 

The inclusion of the lexicons results in higher coverage, which might be part of the explanation for
the improvement of DsDs. The question remains whether the tagger learns to rely only on this additional signal, or it generalizes beyond it. Therefore, we first turn to inspecting out-of-vocabulary (OOV) items. OOV items are the  key challenge in part-of-speech tagging, i.e., to correctly tag tokens unseen in the training data. 
  
In Figure~\ref{fig:oov-properties}, we analyze accuracy on different groups of tokens: The \emph{in lex+train} tokens that were seen both in the lexicon and the training data, the \emph{in train only} tokens seen in the training data but not present in the lexicon, the \emph{in lex only} tokens that were present in the lexicon but not seen in the training data and the \emph{true OOV} tokens that were neither seen in training nor present in the lexicon. 
Figure~\ref{fig:oov-properties} (b) shows means over the 21 languages, Figure~\ref{fig:oov-properties} (c) provides details per language. The first take-away is that in many cases the tagger does learn to use information beyond the coverage of the lexicon. 
The embedded knowledge helps the tagger to improve on tokens which are in train only (and are thus not in the lexicon, green bars). For true OOVs (orange bars), this is the case for some languages as well Figure~\ref{fig:oov-properties} (c), i.e.,  improvements on true OOVs can be observed for Bulgarian, German, Greek, English, Finish, Croatian, Italian and Portuguese. Over all 21 languages there is a slight drop on true OOVs: -0.08, but this is a mean over all languages, for which results vary, making it important to look beyond the aggregate level. 
Over all languages except for Hungarian, the tagger, unsurprisingly, improves over tokens which are both in the lexicon and in the training data (see further discussion in Section~\ref{sec:discussion}).

\subsection{Results on the test sets} 

\begin{table*}[ht!]
\caption{Comparison of our best model to prior work. Best result in boldface. } 
\centering
\resizebox{\textwidth}{!}{
\begin{oldtabular}{l|rrrrrr}
\toprule
&  \multicolumn{6}{c}{\textsc{Test sets}} \\
\textsc{Language} &  \textsc{Das} & \textsc{Li} & \textsc{Garrette} & \textsc{Agi{\' c}} &  \textsc{T\"{a}ckstr\"{o}m} & \textsc{DsDs} \\
\midrule
Bulgarian (bg) & -- & -- & 83.1 & 77.7 & \textbf{87.8} & 84.0 \\
Croatian (hr) &  -- &  -- &  -- & 67.1 & -- & \textbf{79.5} \\
Czech (cs) &  -- &  -- &  -- & 73.3 & 80.3 & \textbf{86.8} \\
Danish (da) & 83.2 & 83.3 & 78.8 & 79.0 & \textbf{88.2} & 84.7\\
Dutch (nl) &  79.5 & 86.3  & -- &  -- & \textbf{85.9} & 83.8\\
English (en) &   -- & \textbf{87.1} &  80.8 & 73.0 & -- & 85.4\\
French (fr) &   -- &  -- & 85.5 & 76.6 & 87.2 & \textbf{88.9}\\
German (de) & 82.8 & 85.8 & 87.1 & 80.2 & \textbf{90.5} & 84.4\\
Greek (el) & 82.5 & 79.2 & 64.4 & 52.3 & \textbf{89.5} & 80.9 \\
Hindi (hi) &  -- &  -- &  -- & \textbf{67.6} & -- & 62.6\\
Hungarian (hu) &  -- &  -- & 77.9 & 70.4 & -- & \textbf{78.1} \\
Italian (it) &  86.8 & 86.5 & 83.5 & 76.9 & 89.3 & \textbf{92.2} \\
Norwegian (no) &  -- &  -- & 84.3 & 76.7 & -- & \textbf{86.3}\\
Persian (fa) &  -- &  -- &  -- & \textbf{59.6} & -- & 43.5\\ 
Polish (pl) & -- &  -- &  -- & 75.1 & -- & \textbf{84.6} \\
Portuguese & 87.9 & 84.5 & 87.3 & 83.8 & \textbf{91.0} & 89.8 \\
Spanish (es) & 84.2 & 86.4 & 88.7 & 81.4 & 87.1 & \textbf{91.3}\\
Swedish (sv) & 80.5 & 86.1 & 76.1 & 75.2 & \textbf{88.9} & 82.7\\
\midrule
\textsc{Avg(8: Das)} & 83.4 & 84.8 & -- & -- & \textbf{88.8} & 86.2\\
\textsc{Avg(12: Garrette)} & -- & -- & 81.4 & 75.3 & -- &  \textbf{85.7}\\
\textsc{Avg(10: Agic $\cap$ T\"{a}ckstr\"{o}m)} & --& -- &  -- & 75.6 & \textbf{87.9} & 86.6\\
\bottomrule
\end{oldtabular}
}

\label{tbl:testsets}
\end{table*}

On the test sets presented in Table~\ref{tbl:testsets}, \textsc{DsDs} reaches
86.2 over 8  languages of Das and
Petrov (2011), compared to their 83.4 and the wikily-based approach which reaches 84.8. This shows
that our ``soft'' inclusion of noisy dictionaries
is superior to hard-inclusion based approaches. 

On the 12 languages by Garrette \& Baldridge (2013), \textsc{DsDs}  reaches an accuracy of 85.7, outperforming prior approaches (Garrette: 81.4, Agic: 75.3).
The approach by T\"{a}ckstr\"{o}m et al., (2013) reaches higher overall accuracy than \textsc{DsDs} on the 8 languages by Das and Petrov (2011) (86.2 vs 88.8). 
Their approach relies on 1-5M near-parallel instances from Europarl, which lacks the broad coverage from WTC and is cleaner data. Our model is trained on only 5k noisy instances, and benefits from the diverse additional sources of supervision (pre-trained embeddings and dictionaries).  \textsc{DsDs} even outperforms their approach on four language: Czech, French, Italian and Spanish.

\subsubsection{Results on more languages} 

All data sources employed in our experiment so far are considerably high-resource. However, for true low-resource languages, we cannot safely assume the availability of all disparate information sources. Table~\ref{tbl:closelowlang} presents results for four additional languages where some supervision sources are missing. There are several take-aways.

First, adding lexicon information always helps, even in cases where only 1k entries are available (Basque). Second, for closely-related languages such as Serbian and Croatian, using resources for one aids tagging the other. In absence of a Serbian dictionary, using the Croatian resource aids tagging for Serbian. Moreover, more modern resources are a better fit: the Croatian WTC projections result in better training data than the in-language Serbian Bible projections, which is older language text and its OOV rate is  higher. In absence of further resources, our ``take what you can get'' approach is beneficial. For instance, we only had a morphological dictionary for Estonian, and only off-the-shelf embeddings for Tamil, and using those improves tagging accuracy.

\begin{table}[h]
\caption{Results for languages with missing data sources: WTC projections, Wiktionary (W), or UniMorph (U). Test sets ({\sc Test}), projection sources ({\sc Proj}), and embeddings languages ({\sc Emb}) are indicated. Comparison to TnT (Brants, 2000) trained on {\sc Proj}.}
\resizebox{0.95\textwidth}{!}{
\begin{oldtabular}{lccccc|cccc}
\toprule
                     & & &   & \multicolumn{2}{c|}{\textsc{Lex} ($10^3$)} & \multicolumn{4}{c}{{\sc Test sets}}\\
\textsc{Language} & \textsc{Test} & \textsc{Proj} & \textsc{Emb} & W & U & TnT & 5k & TC$_W$ & \textsc{DsDs}  \\
\midrule
Basque (eu) & UD & Bible & eu & 1 & -- &  57.5 &  60.4 & 60.4 & \textbf{61.5}\\
Basque (eu) & CoNLL & Bible & eu & 1 & -- & 57.0 & 59.7 & 59.7 & \textbf{61.2} \\
Estonian (et) & UD & WTC & et & -- & 10 &  79.5 & 81.5 & -- & \textbf{81.8}\\
Serbian (sr) & UD & WTC (hr) & hr & (hr) 20 & -- & 84.0 & 85.1 & 85.7 & \textbf{86.1} \\ 
Serbian (sr) & UD & Bible (sr) & hr  & (hr) 20 & --& 77.1 & 82.3 & 82.6 & \textbf{83.4}\\
Tamil (ta) & UD & WTC & ta &  -- & -- & 58.2 & \textbf{60.0} & -- & -- \\
\bottomrule
\end{oldtabular}
}
\label{tbl:closelowlang}
\end{table}

\section{Discussion}
\label{sec:discussion}

The inclusion of lexicons results in
higher coverage. However, even a very small dictionary can help, and as indicated above, the positive effect of lexical information is already observable extrinsically,  in the improved OOV tagging accuracy.  Here we dig deeper into the effect of including lexical information by a) examining learning curves with increasing dictionary sizes, b) relating tag set properties to performance, and finally c) having a closer look at model internal representations, by comparing them to the representations of the base model that does not include lexical information. We hypothesize that when learning from dictionary-level supervision,  information is propagated through the representation layers so as to generalize beyond simply relying on  the respective external resources.  We introspect the model-internal representations in two ways: through visualizations and by inspecting them via probing tasks.

\subsection{Learning curves}\label{sec:sampling}
The lexicons we use so far  are of  different size (Table~\ref{tbl:results}), spanning from just 1,000 entries to considerable dictionaries of several hundred thousands entries. In a low-resource setup, large dictionaries might not be available. It is thus interesting to examine how tagging accuracy is affected in case of less dictionary supervision. We examine two cases for \textsc{DsDs}: randomly sampling dictionary entries and sampling by word frequency, both for increasing dictionary sizes: 50, 100, 200, 400, 800, 1600 word types. The latter is motivated by the fact that an informed dictionary creation (under limited resources) might be more beneficial. We estimate word frequency by using the UD training data sets (which are otherwise not used).

\begin{figure}[th!]
  \subfloat[a][Average effect over 21 languages of high-freq and random dictionaries]{            
    \includegraphics[width=.5\textwidth]{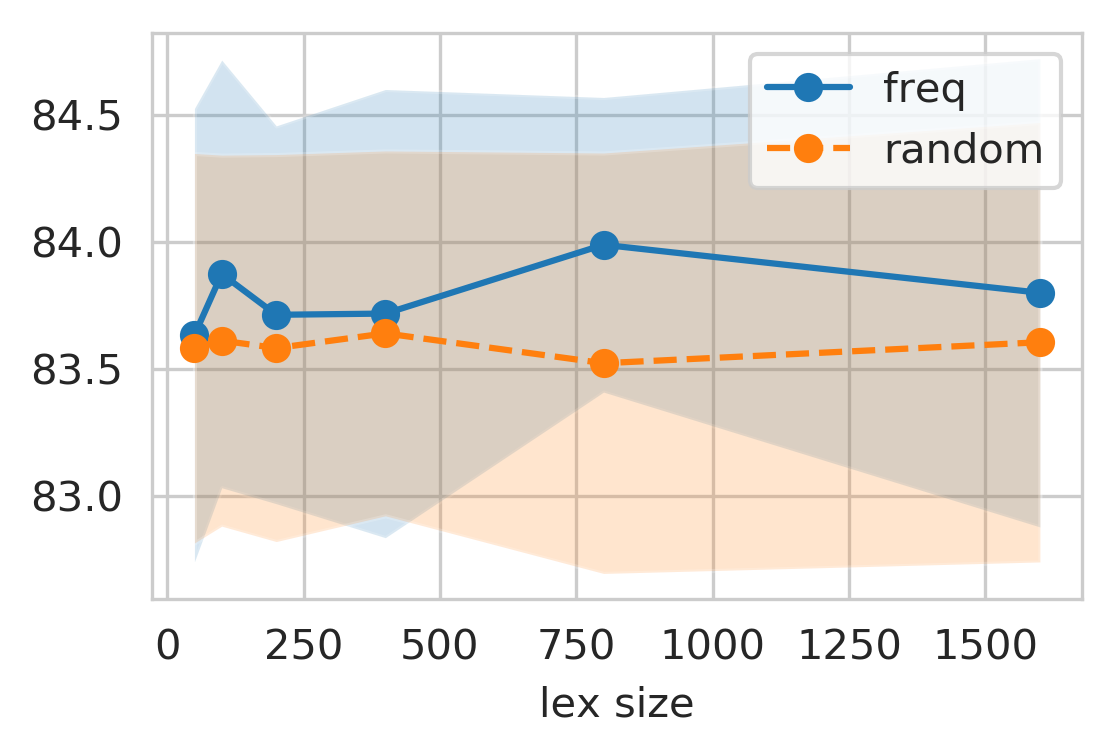}}
  \subfloat[b][Effect for subset of language families of high-freq and random dictionaries]{
    \includegraphics[width=.6\textwidth]{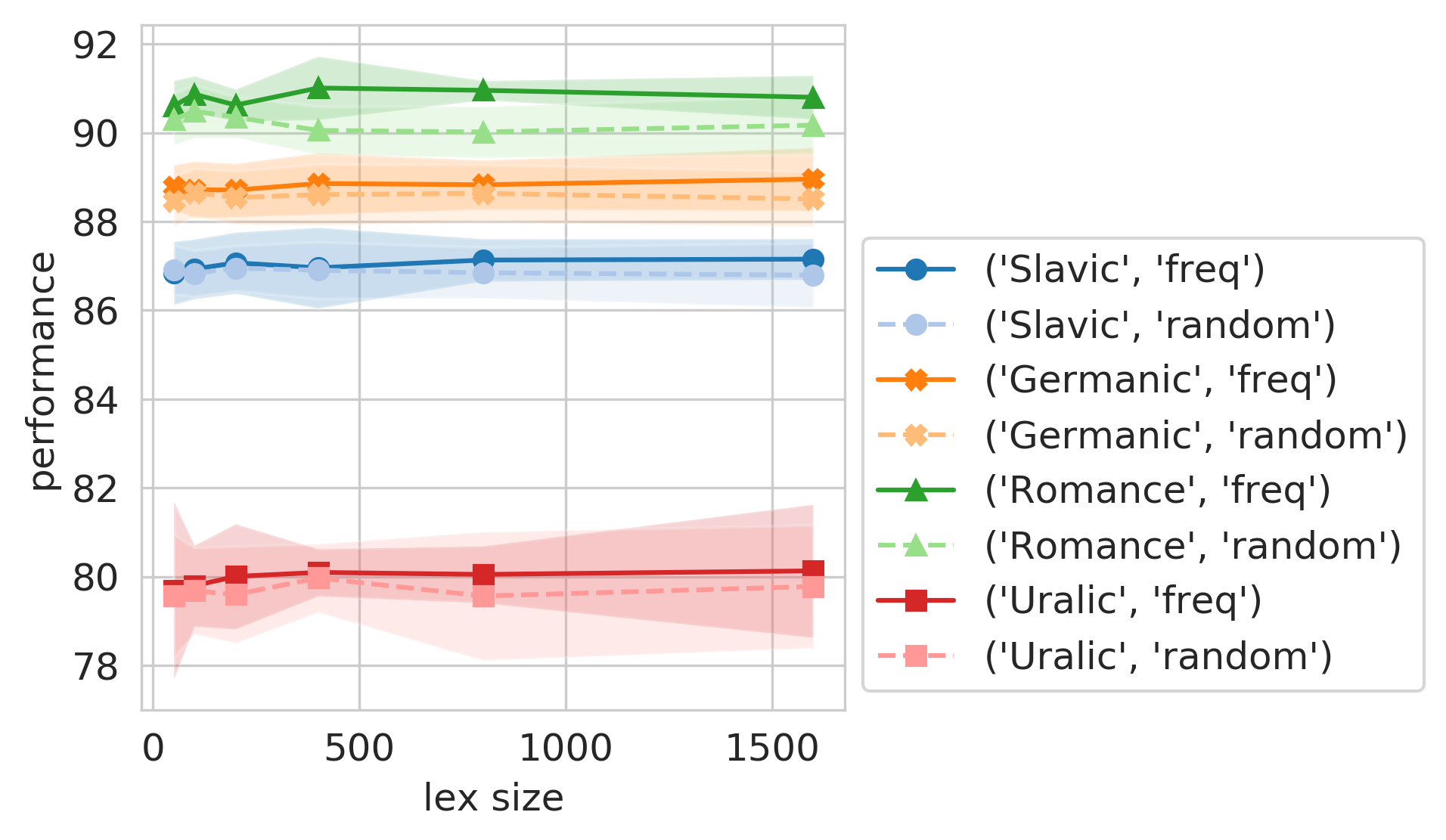}}
      \qquad
    \subfloat[c][Danish: language with clear improvement from high-freq dictionary (base: 89.8, all: 90.1)]{            
    \includegraphics[width=.5\textwidth]{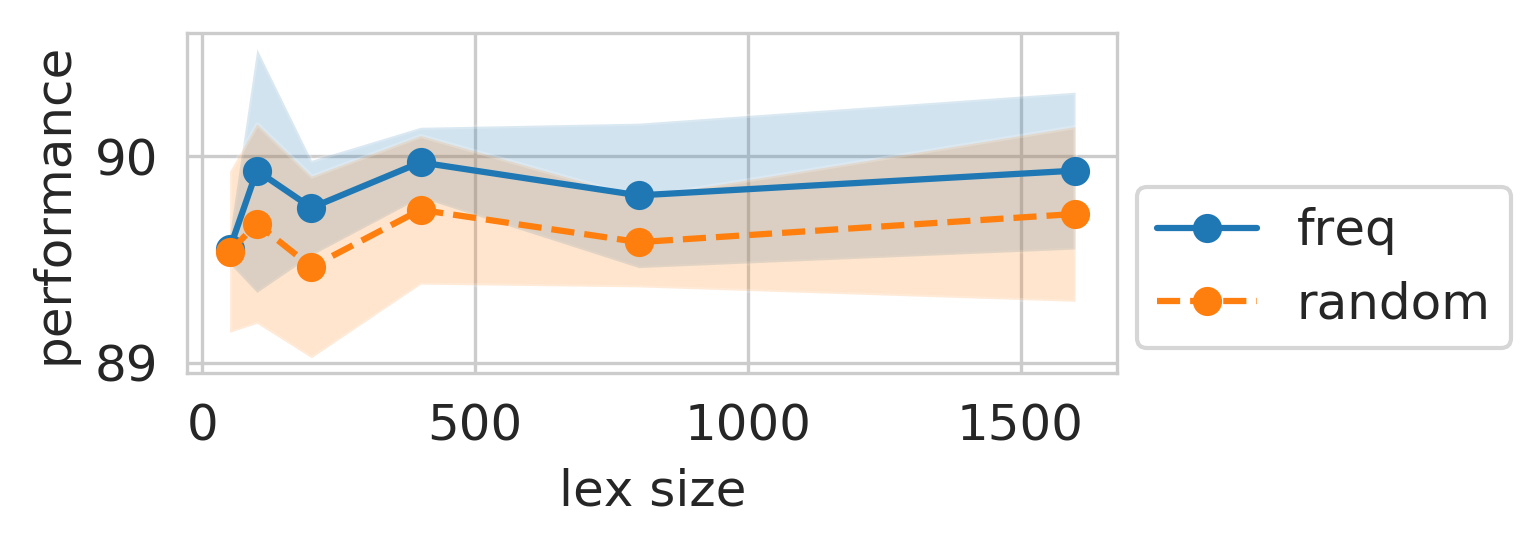}} 
  \subfloat[d][Italian: language with clear positive effect of high-freq dictionary (base: 92.2, all: 93.7)]{            
        \includegraphics[width=.5\textwidth]{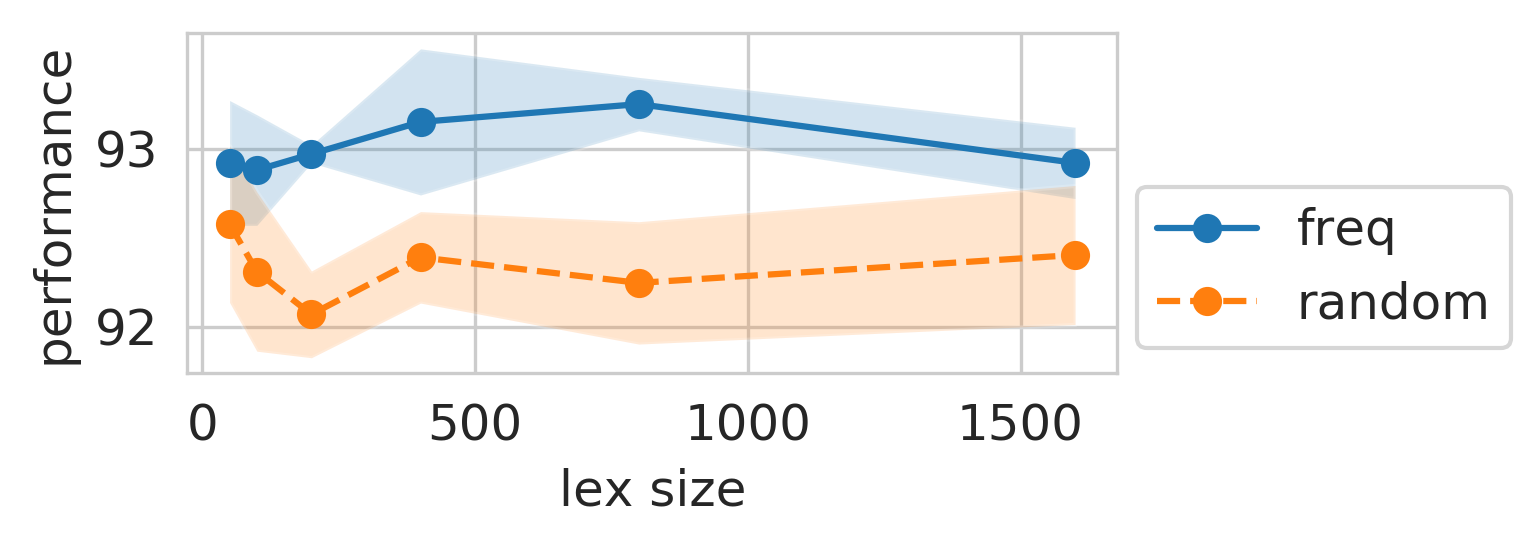}}
    \qquad
    \subfloat[e][Hungarian: language with observed negative effect (base: 77.5, all: 76.2)]{            
    \includegraphics[width=.5\textwidth]{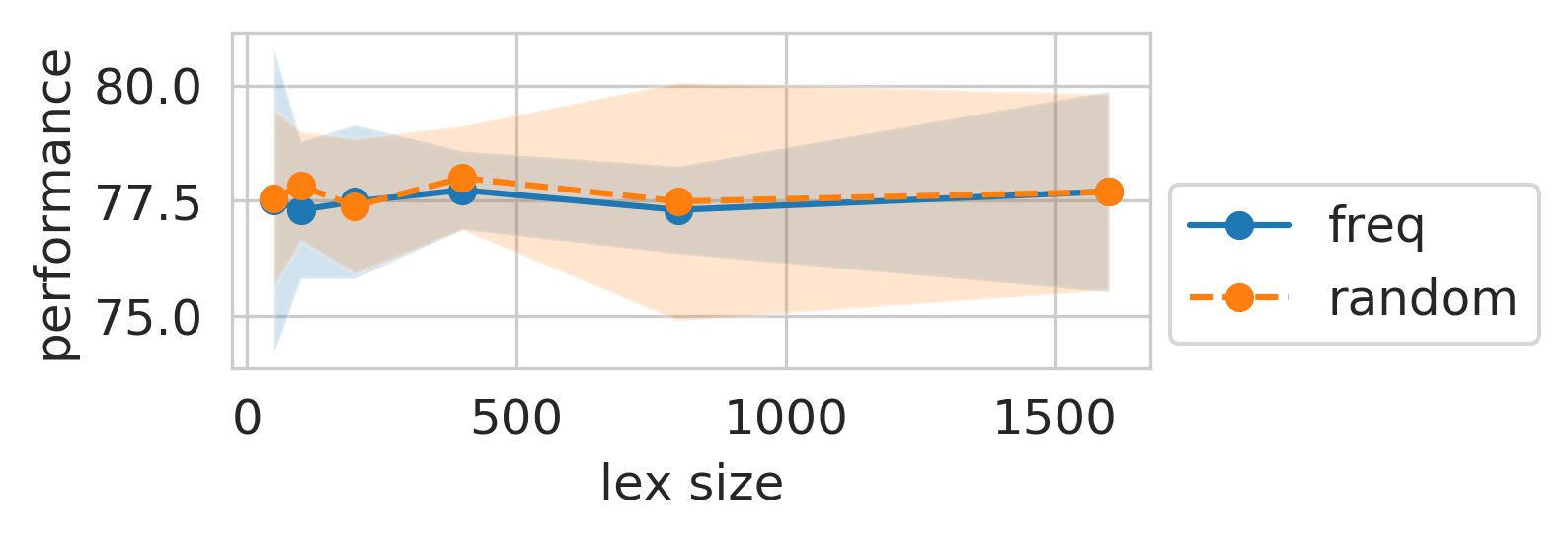}} 
    \subfloat[f][Czech: language with observed negative effect (base: 87.5, all: 87.4)]{            
    \includegraphics[width=.5\textwidth]{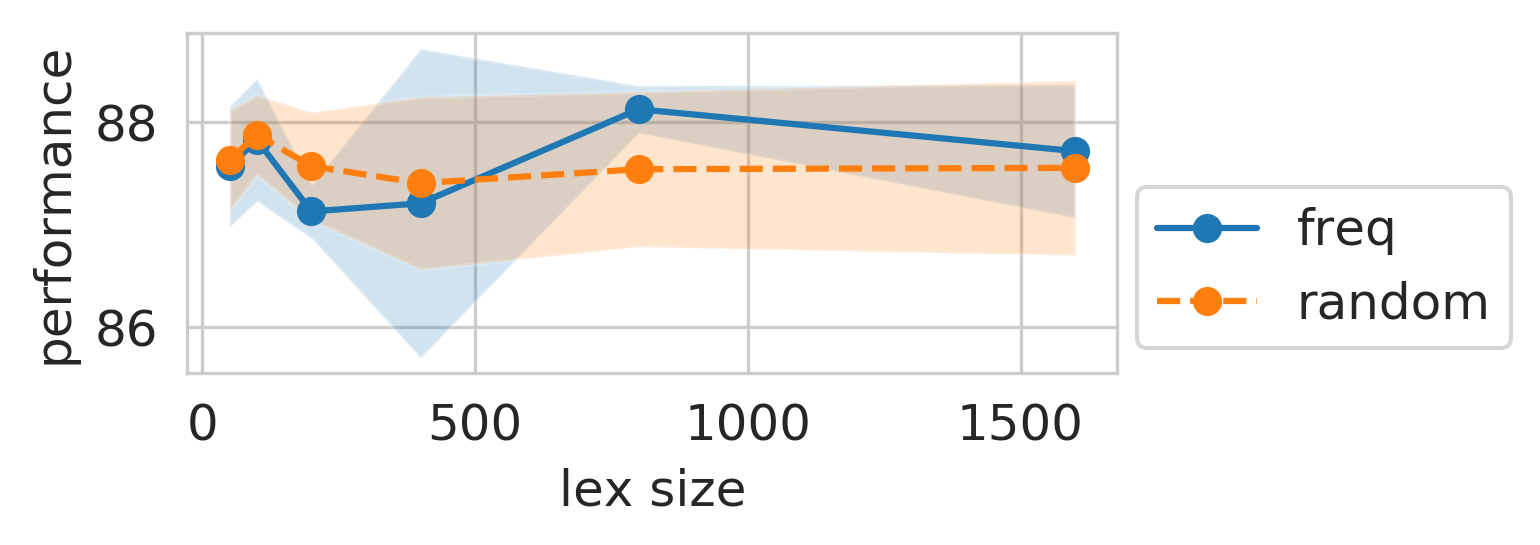}}. 
\caption{Learning curves over increased dictionary sizes.}
    \label{fig:curves}
\end{figure}

Figure~\ref{fig:curves} presents the results of training \textsc{DsDs} with increasing dictionary sizes. Figure~\ref{fig:curves} (a) provides means over the 21 languages (with confidence intervals of $\pm 1$ standard deviation based on three runs). We note that sampling by frequency is overall more beneficial than random sampling. The biggest effect of sampling by frequency is observed for the Romance language family, see Figure~\ref{fig:curves} (b). 

Consider the plots for selected languages in Figure~\ref{fig:curves} (c-f). What is interesting to note is that more dictionary data is not always necessarily beneficial. Sometimes a small but high-frequency dictionary approximates the entire dictionary well. This is for instance the case for Danish,  Figure~\ref{fig:curves} (c), where sampling by frequency approximates the entire dictionary well (`all' achieves 90.1, while using 100 most frequent entries is close: 89.93). Frequency sampling also helps clearly for Italian, Figure~\ref{fig:curves} (d), but here having the entire dictionary results in the overall highest performance. 

For some languages, the inclusion of lexical information does not help, not even at smaller dictionary sizes. This is the case for Hungarian, French and Czech. The Hungarian learning curve shown in Figure~\ref{fig:curves} (e) is essentially flat, and training with the entire dictionary (`all') drops performance below the baseline. For Czech, this is less pronounced, as the performance stays around baseline. Relating these negative effects to the results from the tag set agreement analysis (Figure~\ref{fig:tagset-typelevel-all}), we note that Hungarian is the language with the largest \textit{disjoint} tag set. Albeit the coverage for Hungarian is good (around .5), including too much contradictory tag information has a clear deteriorating effect. Consequently, neither sampling strategy works and the learning curves are similar. Czech, which has smaller coverage, sees a negative effect as well, as half of the dictionary entries have disjoint tag sets. In contrast, Italian is the language with the highest dictionary coverage \textit{and} the highest proportion of equal tag sets, thereby providing a large positive benefit.

We conclude that when dictionaries are not available, creating them by targeting high-frequency items is a pragmatic and valuable strategy. At times a small dictionary, which does not contain too contradictory tag sets, can be beneficial when used with \textsc{DsDs}.

\subsection{Dictionary coverage and tag set agreement}~\label{sec:wiktionary_analysis}

\begin{figure}[]
\subfloat[][Proportion of tokens unseen in the training data, in the lexicon or in both (true OOV's). Lighter bars are proportion of correctly labeled portion, dark bars are proportion of errors. Averaged over three runs.]
  {
  \includegraphics[width=\textwidth]{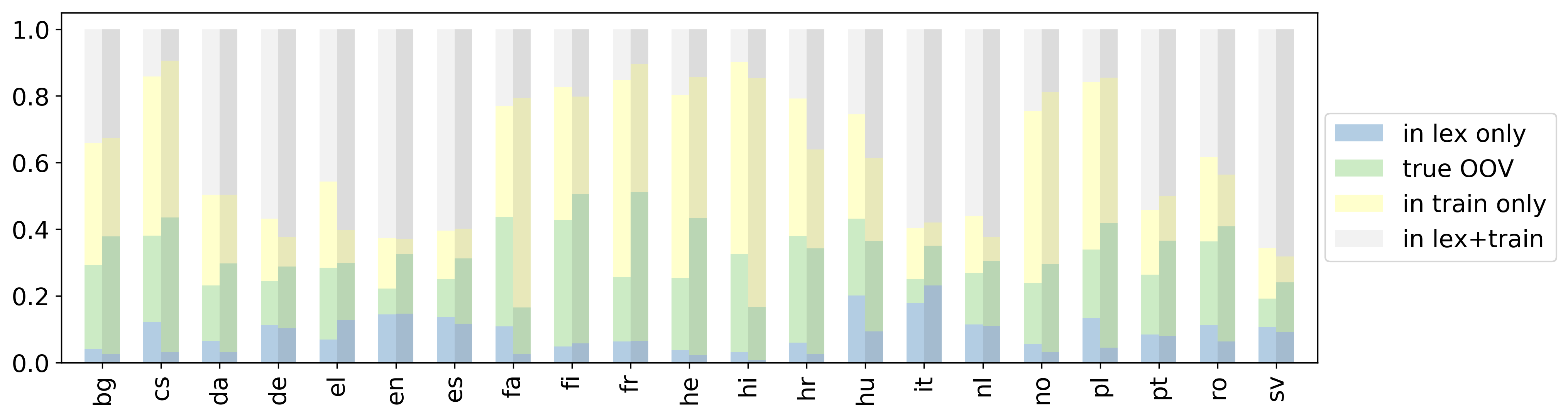}}
  \\
\subfloat[][Development set size and size of correctly and incorrectly labeled parts. Averaged over three runs.]
  {
  \includegraphics[width=.4\textwidth]{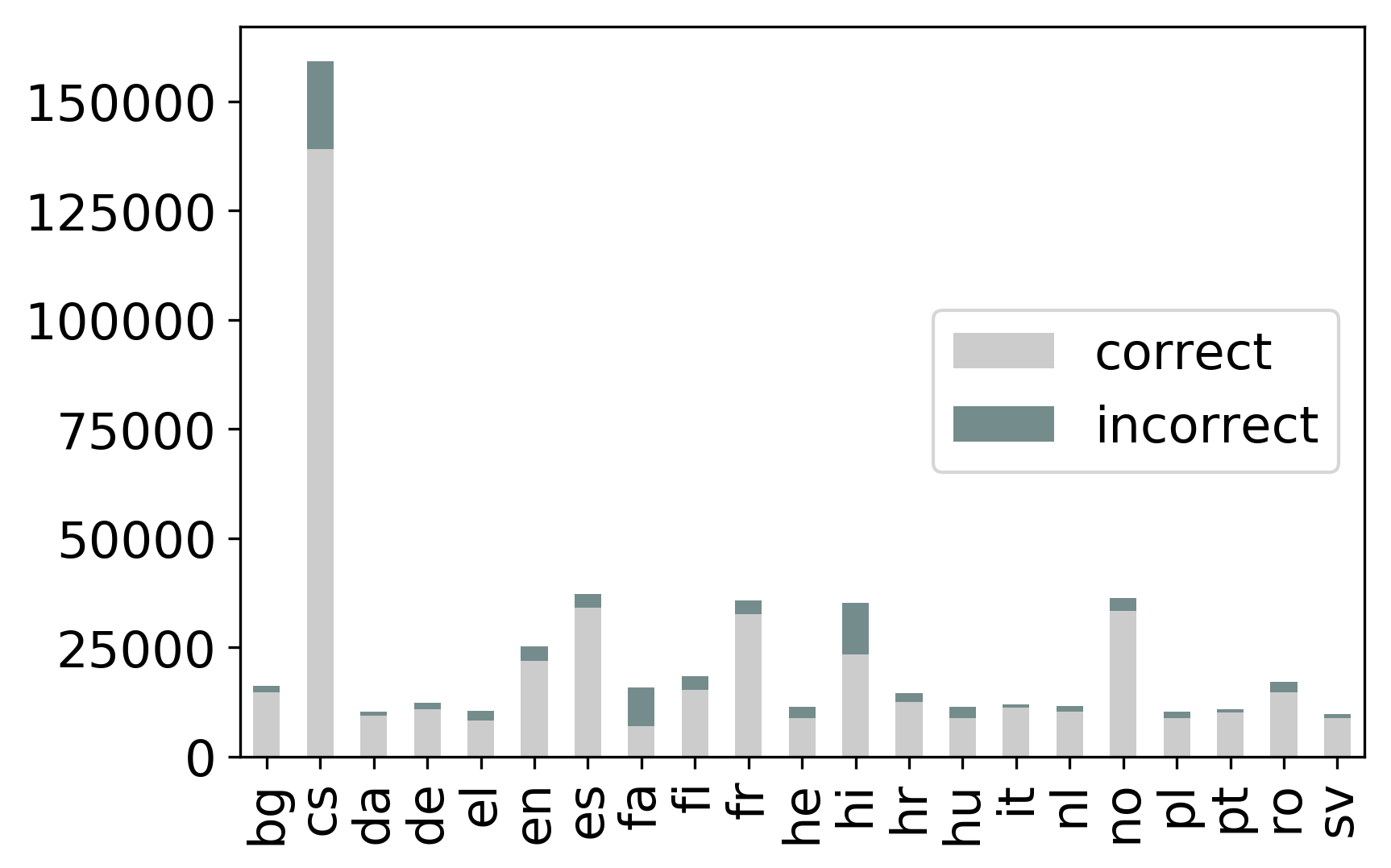}}
\hspace{.5cm}
\subfloat[][Token-wise coverage and tagset agreement between dictionary and gold tags by accuracy]{            
    
    \includegraphics[width=.45\textwidth]{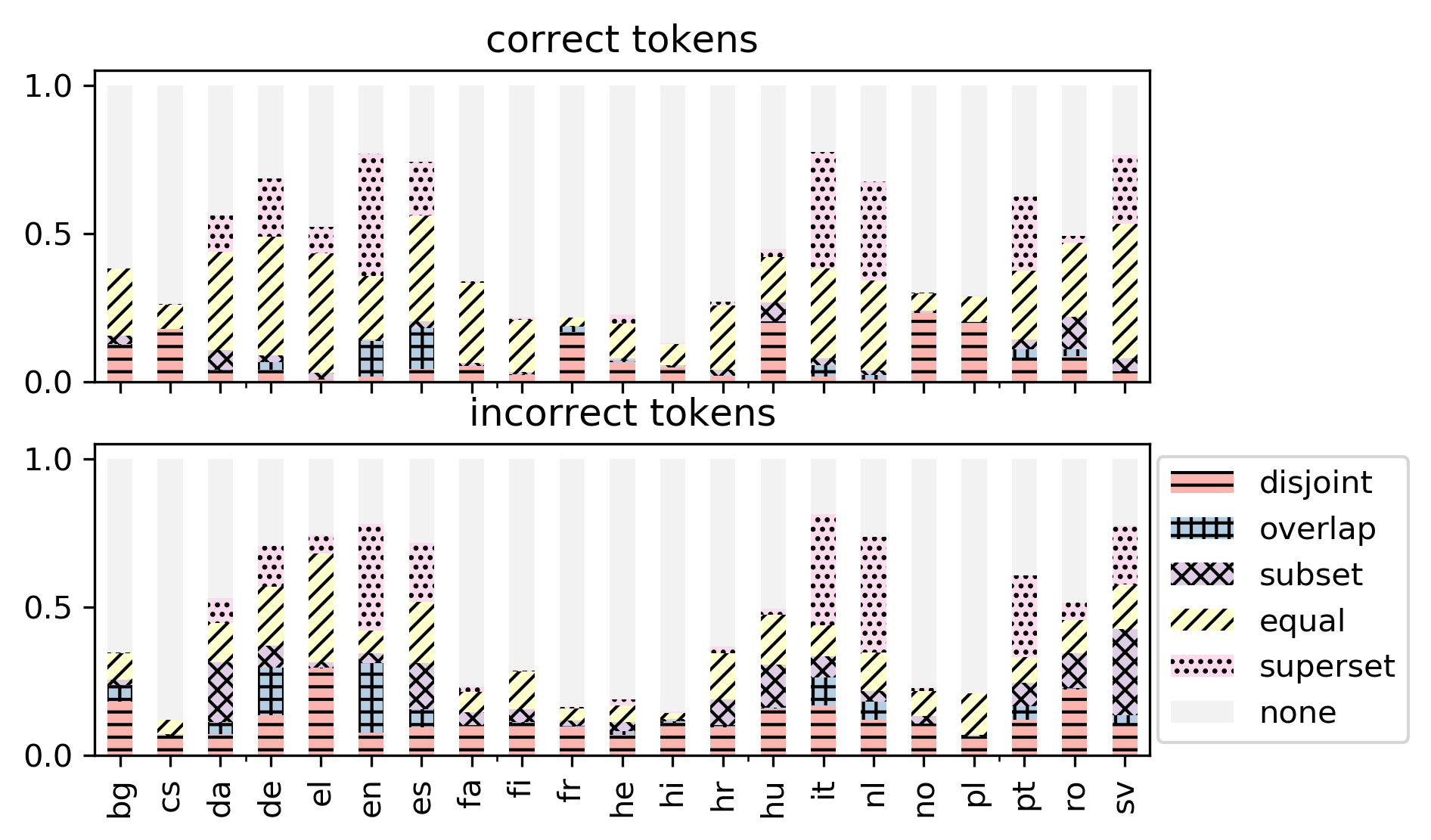}}
    \\
\subfloat[][Type-wise coverage and tagset agreement between dictionary and gold tags by ambiguity]{            
    \includegraphics[width=.45\textwidth]{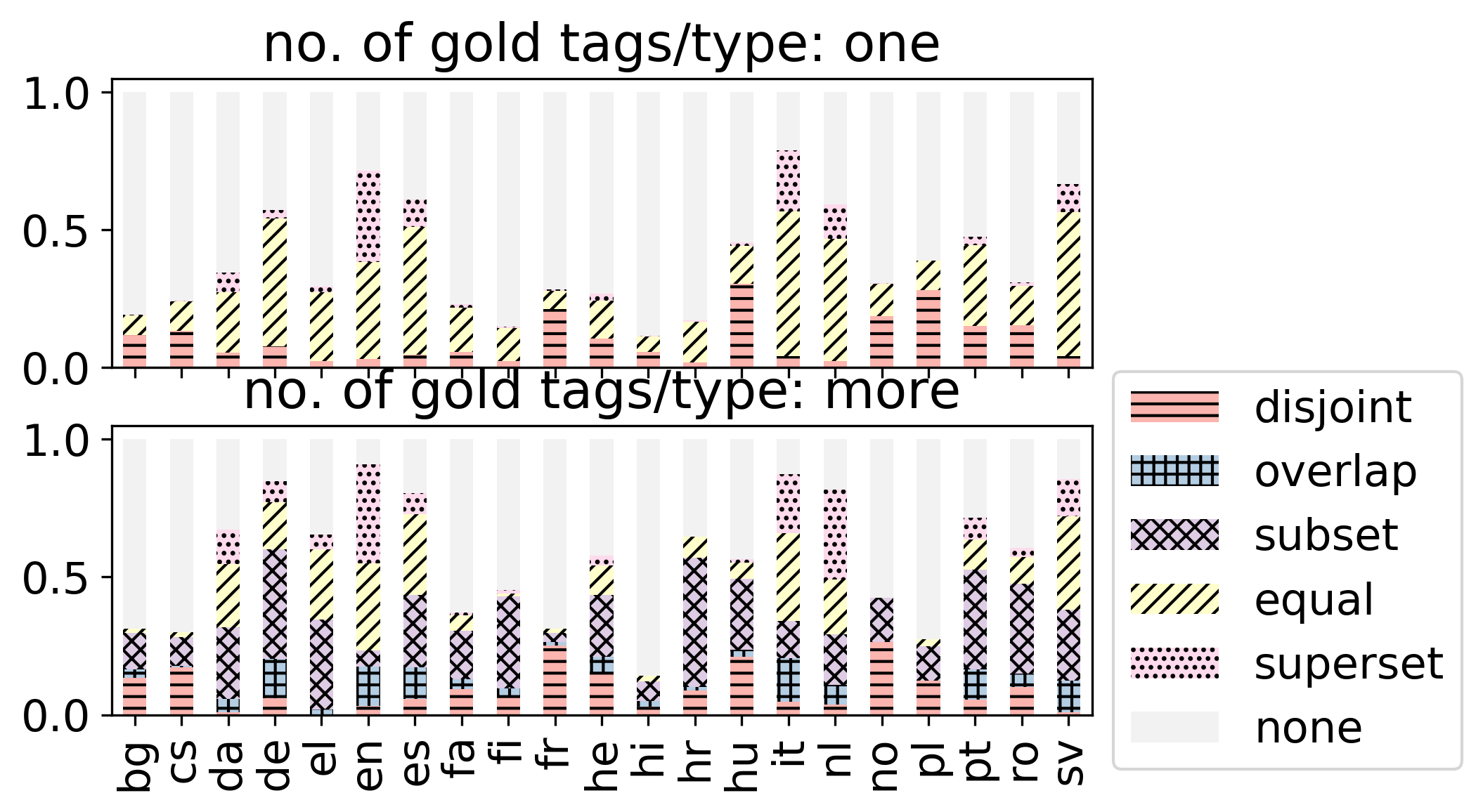}}
\qquad
\subfloat[][Type-wise coverage and tagset agreement between dictionary and gold tags by top and bottom frequent types]{            
    \includegraphics[width=.45\textwidth]{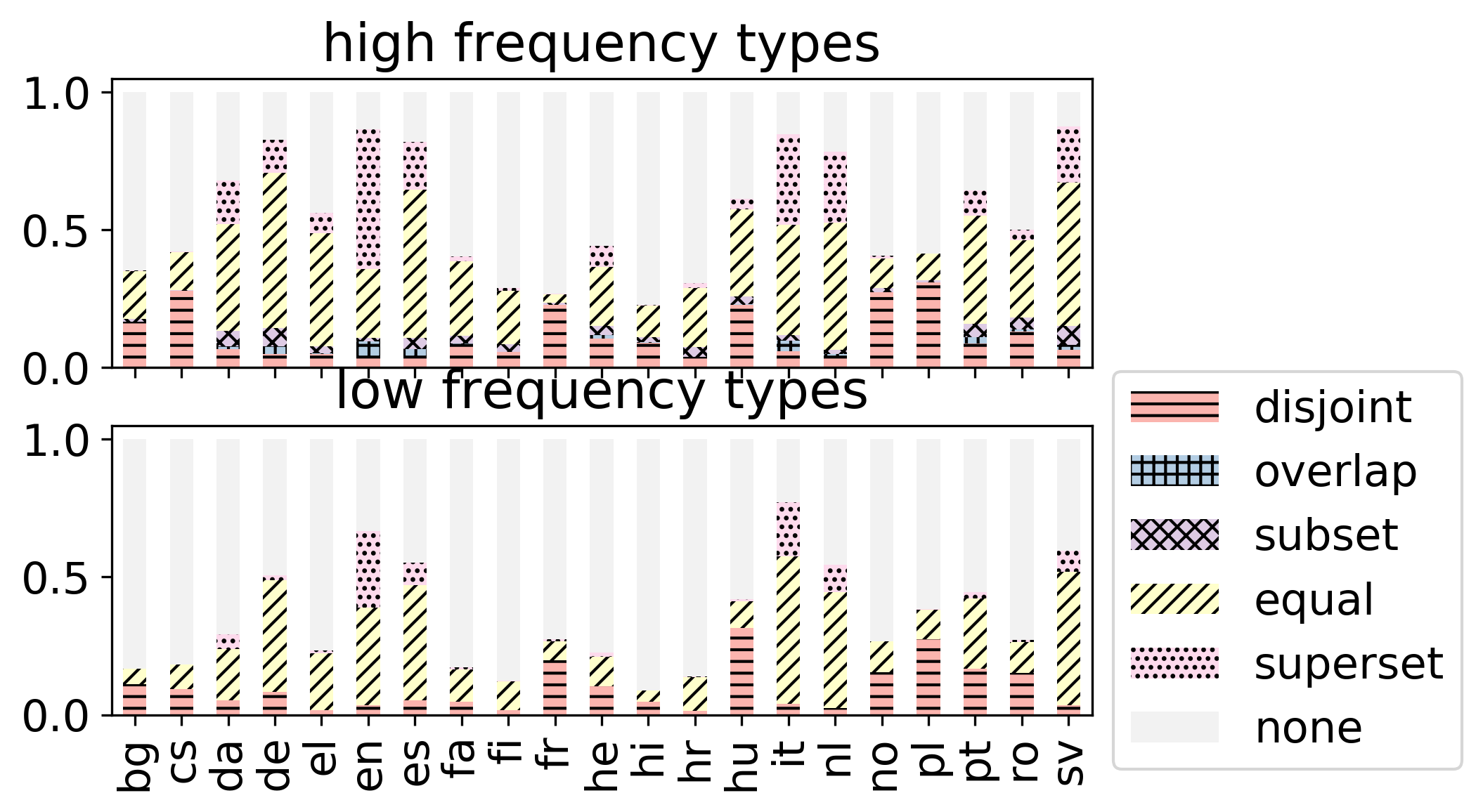}}

   \caption{Coverage and tag set agreement.}
   \label{fig:tagsets}
\end{figure}

In the following we inspect the variation in the lexicon contributions across languages by comparing coverage and tag set agreement on correctly and incorrectly labeled tokens.
Because we are analyzing differences between languages as well as between errors and successes we here abstract away from the underlying sample size variation by comparing proportions. The sizes of the development sets and their correctly and incorrectly labeled parts are included for reference in Figure~\ref{fig:tagsets} (b). The token-level analyses are followed by two type-level analyses which compare tag set agreement with respect to the tag ambiguity observed in the gold labels for the development set, and with respect to word frequency using estimates based on the UD training data as previously noted.

The first analysis inspects the differences in proportions on four subsections of the development set, as introduced above: the \emph{in lex+train} tokens, the 
\emph{in train only} tokens, the \emph{in lex only} tokens and the \emph{true OOVs}. The proportion of these four data subsets in the correctly and the incorrectly labeled tokens are shown side by side in Figure~\ref{fig:tagsets} (a) in lighter and darker shades, respectively. If the OOV-status of a word was unrelated to performance, the lighter and darker bars would be of identical size, this is not the case and we can observe that the true OOVs make up a significantly larger share of the errors than of successes (two-tailed paired Student's t-test: $p=0.007$). Similarly, seen across all languages the shift in the size of the proportion of true OOVs is made up by more correct labeling of a larger proportion of \emph{in train only} (two-tailed paired Student's t-test: $p=0.014$) and \emph{in lex only} (two-tailed paired Student's t-test: $p=0.020$), whereas the proportion of \emph{in lex+train} does not significantly differ between the correctly and incorrectly labeled parts (two-tailed paired Student's t-test: $p=0.200$).\footnote{Significance based on an $\alpha$-level of 0.05}

We can contrast the partition into coverage-based sub-parts with the tag set-agreement sub-partition shown in Figure~\ref{fig:tagsets} (c). In this figure the categories follow the description in Section~\ref{sec:resources} with all tokens not present in the lexicon marked in grey and the remaining tokens split by the agreement between observed gold tags and the tag set recorded in the lexicon (only UD tags are considered). In the upper half of the figure, reporting tag set agreement as proportions of correctly labeled tokens, we observe that five languages with relatively low lexicon coverage, namely Bulgarian (bg), Czech (cs), French (fr), Norwegian (no) and Polish (pl), have large shares of correctly labeled tokens with disjoint tag sets. For the four latter languages and Hungarian (hu), which has a higher but noisy lexicon coverage, the share of correctly labeled tokens with disjoint tag sets is larger, in fact, than the share of incorrectly labeled ones. This can be interpreted as evidence of a strategy which favors the training set information over the lexicon---a strategy which is associated with a lower performance when lexicons are included for Czech, French and Hungarian while not for Bulgarian, Norwegian and Polish. Further, in all languages except French (fr), Norwegian (no) and Polish (pl) the share of tokens with equal tag sets is similar or larger for correctly labeled tokens than for incorrectly labeled ones, whereas the potentially confusing \emph{subset} and \emph{overlapping} tag sets tend to make up a larger share of the incorrectly labeled tokens.

Type-level analysis of the tag set agreement with respect to tag ambiguity and word frequency is shown in Figure~\ref{fig:tagsets} (d) and (e). All words present at most once in the UD data are considered low frequency while the cut-off minimal count for high frequency items is set to include at least 10\% of the development set.\footnote{The count range for the high frequency group cut-off is between 3 and 16} Most notably the large share of disjoint, that is, contradicting information is also more prominently present in word types observed with only one gold tag compared to types observed with multiple gold tags for the three low performing languages also mentioned above (Czech, French, Hungarian). For five of the six most disjointly labeled languages (Bulgarian, Czech, French, Norwegian and Polish), high frequency types are even more likely to present this contradicting information in the lexicon, but this may be an artifact of high frequency types being generally more likely to be in the lexicon as evidenced by the larger proportion of grey bars in Figure~\ref{fig:tagsets} (e).

\subsection{Analysis of character-based word encodings}
The pre-trained word embeddings stay fixed in our model (see Section~\ref{sec:updatingembeds}). However, the character-based word encodings get updated: This holds true both for the {\sc Base} system and for the top-performing {\sc DsDs} tagger.  As a target for assessing the flow of information in the neural tagger, we thus focus on the character-based word encodings.

The word-level is relevant as it is the granularity at which the tagger is evaluated. The word embeddings may already have encoded PoS-relevant information and the lexicon embeddings explicitly encodes PoS-type-level information. By contrast, the character-based word encodings are initialized to be uninformative and any encoding of PoS-related information is necessarily a result of the neural training feedback signal.

For these reasons we employ two different strategies that enable us to query the character-based word representations of the tagger in order to compare variation across languages as well as between the base tagger and the \textsc{DsDs} lexicon-enriched architecture. 

\subsubsection{Probing models}

\begin{figure}[h!]
   \includegraphics[width=\textwidth]{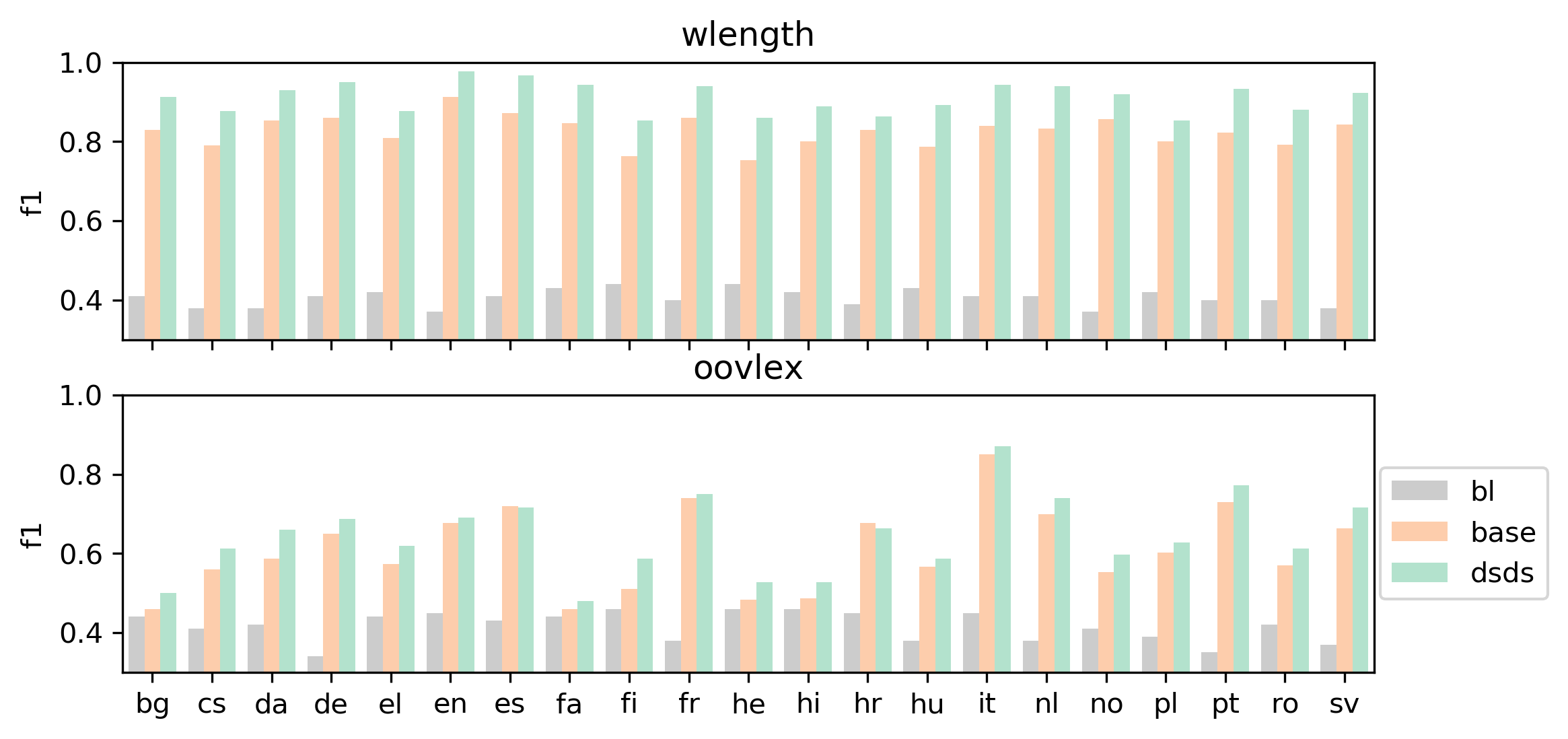}
   \caption{Macro F1 scores for stand-alone classifiers on the probing tasks of predicting which words are long and which are in the lexicon, respectively. The baseline (bl) is a simple majority baseline. The base- and DsDs-informed classifiers were trained on character-based word representations from the neural taggers with and without access to lexical information, respectively.}
   \label{fig:probing}
 \end{figure}

Probing tasks, or diagnostic classifiers, are separate classifiers which use representations extracted from any facet of a trained neural model as input for solving a separate task.
Following the intuition of~\cite{adi:ea:2017}, if the target can be predicted, then the information must be encoded in the representation. However, the contrary does not necessarily hold: if the model fails it does not necessarily follow that the information is not encoded, as opposed to not being encoded in a useful way for a probing task classifier.

As the internal representations stored in neural models are not immediately interpretable, probing tasks serve as a way of querying neural representations for interpretable information. The probing task objective and training data is designed to model the query of interest. The representation layer we query in this work is the word-level output from the character embedding sub-model. This part of the word-level representation starts out uninformative and thus without prior prediction power on the classifier objectives.

Following Shi et al.~\shortcite{shi:ea:2016} and Gulordava et al~\shortcite{gulordava:ea:2018}, we use a logistic regression classifier setup and a constant input dimensionality of 64 across tasks~\cite{conneau:ea:2018}. The classifiers are trained using 10-fold cross-validation for each of three trained runs of each neural model and averaged. We include a majority baseline and report macro F1-scores, as we are dealing with imbalanced classes. The training vocabulary of both probing tasks is restricted to the neural tagger training vocabulary, that is, all word types in the projected training data, as these are the representations which have been subject to updates during training of the neural model. Using the projected data has the advantage that the vocabulary is similar across languages as the data comes from the same domain (Watchtower).

We employ two probing tasks: predicting which words are long, i.e. contain more than 7 characters, and predicting which words are in the lexicon. The word length task is included as a task which can be learned independently of whether lexicon information is available to the neural model. Considering words of 7 characters or more to be long is based on the threshold that was experimentally tuned in the design of the readability metric LIX \cite{bjornsson1983readability}. This threshold also aligns well with the visual perceptual span within which proficient readers from grade four and up can be expected to automatically decode a word in a single fixation \cite{sperlich2015perceptual}.
The task of predicting which words are in the lexicon is designed to probe for the added attention paid to the lexicon by the neural model.

The results on the word length probing task shown on the top half of Figure~\ref{fig:probing} confirm that information relevant to distinguishing word length is being encoded in the neural representation, as expected. However, it is intriguing that the lexicon-informed \textsc{DsDs} representation encodes this information even at at higher degree for the diagnostic classifier. 

On the task of classifying which words are in the lexicon, all neural representations beat the majority baseline, but we also see that this task is harder, given the higher variance across languages. With Spanish (es) and Croatian (hr) as the only exceptions, the DsDs-based representations are generally encoding more of the information relevant to distinguishing which words are in the lexicon, confirming our intuitions that the internal representations were altered.
Note, however, that even the base-tagger is able to solve this task above chance level. This is likely an artifact of how lexicons grow where it would be likely for several inflections of the same word to be added collectively to the lexicon at once, and since the character representations can be expected to produce more similar representations of words derived from the same lemma the classifier will be able to generalize and perform above chance level without the base-model representations having ever been exposed to the lexical resource.

Finally, it is noteworthy that the French lexicon is particularly easy for the classifier to recognize compared to other languages with similarly low lexicon coverage, such as Hebrew (he) and Persian (fa) which both have substantially higher baseline performance (compare to Figure~\ref{fig:tagset-typelevel-all}). This could be an indication of a strong orthographic bias in the content of the French lexicon.

\subsubsection{Similarity within and across PoS}

\begin{figure}
\subfloat[][The difference between cosine distances in {\sc Base} and {\sc DsDs} within PoS set and across PoS sets for {\sc adj, noun, verb}. Dashed lines split the distributions into quartiles. {\sc DsDs} visibly brings words with the same PoS closer together, and draws different PoS further apart.]{
\includegraphics[width=0.7\textwidth]{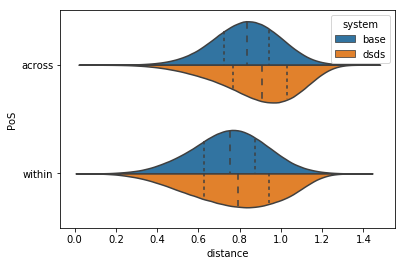}}\\
\subfloat[][Distance distributions within {\sc verb} split by language and system. The distributions by {\sc DsDs} clearly gravitate to left or towards flatter spread, i.e., to increase homogeneity within {\sc verb}.]{
\includegraphics[width=\textwidth]{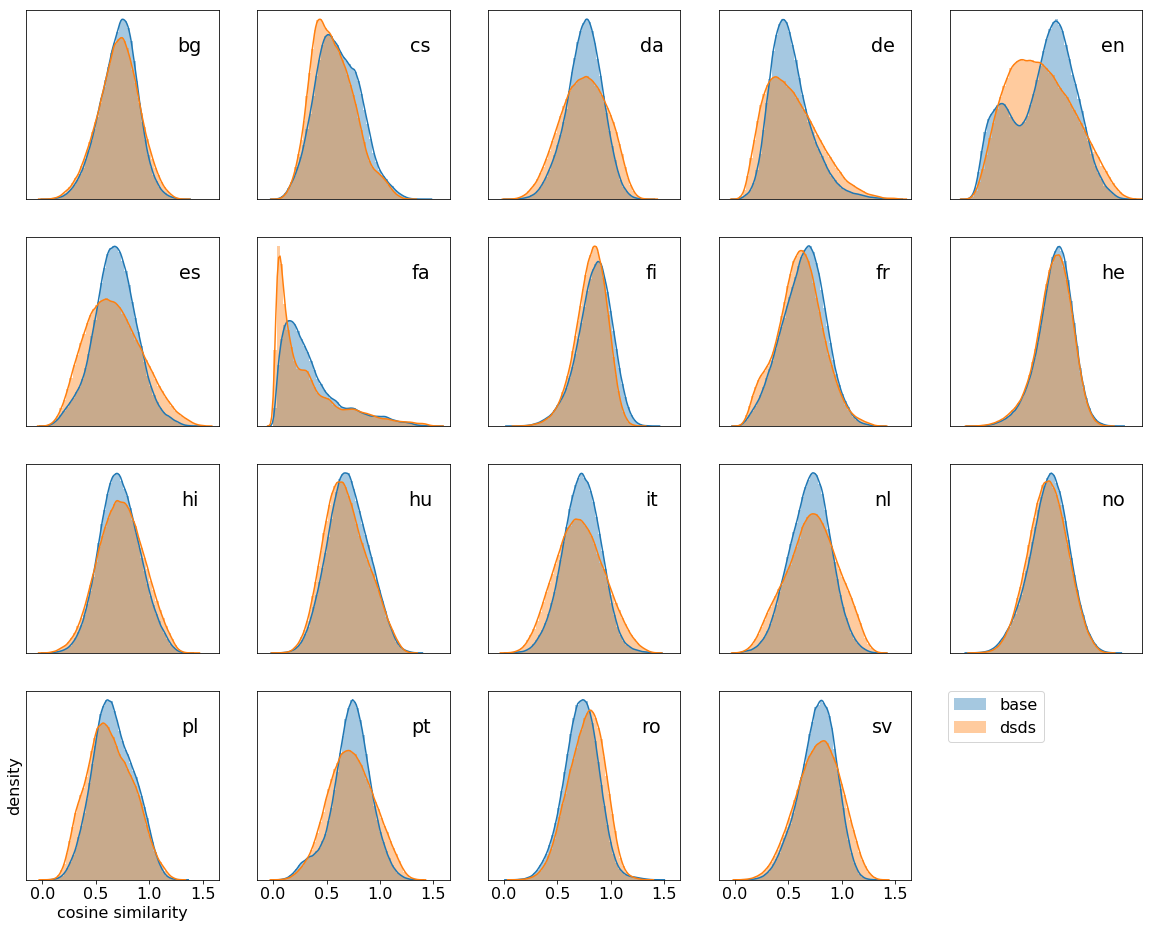}}
\caption{Cosine distances in character-based word encodings in our tagger. Comparison across 19 languages (for which we have Wiktionaries), two PoS taggers {\sc Base} and {\sc DsDs}, averages over three runs, for adjectives, nouns, and verbs. Illustrated in two ways: a) overall, and b) split by language.}
\label{fig:cosineembs}
\end{figure}

In our low-resource setup {\sc DsDs} consistently outperforms the base tagger. We hypothesize that the learning process for {\sc DsDs} thus alters the initial character-based word representations in a more favorable direction for the tagging performance. For this reason, we perform a simple experiment whereby we gauge the final word representations of {\sc Base} and {\sc DsDs} taggers, looking into distances between word vectors by part of speech.

Specifically, for every language, we sample three sets of words---adjectives, nouns, and verbs---each set containing 1,000 words. These words are present in the initial word embeddings, and their parts of speech are known through a dictionary. Then, for each pair of words, we calculate the cosine distances between their word vectors, and plot them separated by language and tagger type. 

Most importantly, the comparison is split for distances within a PoS (adjective to adjective, noun to noun, and verb to verb), and across PoS (adjective to noun or verb, and so on). The hypothesis is that, on average, the training procedure of the higher-performing system would alter the initial word embeddings in a way that brings the embeddings closer together within a PoS, while moving the embeddings further apart across PoS boundaries.

Figure~\ref{fig:cosineembs} illustrates the results: first overall, and then for comparisons within the {\sc verb} set. The plots show a clear trend of {\sc DsDs} to shift left or towards more similarity for comparisons within a PoS, while they shift towards more distance across PoS boundaries. This is by contrast to the {\sc Base} system where we observe a smaller shift effect, which is visually indicated in the plots. All differences between the two systems are statistically significant at $p < 0.01$ via~\cite{mann1947test} across all languages and for both comparisons, within and across PoS. This confirms our hypothesis that the learning process of \textsc{DsDs} alters the representations into more favorable lexically-informed representations.

Note that each density plot in Figure~\ref{fig:cosineembs}b displays a bit under one million vector comparisons. For example, the English {\sc verb} plot for {\sc DsDs} shows $1000$ verbs $\times$ 999 remaining verbs = 999,000 comparisons. The same applies for comparisons across POS, where the other words come from the other PoS groups, but the calculation remains the same. The plot in Figure~\ref{fig:cosineembs} (a) and our significance tests include all three PoS tag sets (adjective, noun, verb) and thus the number of comparisons there is three times larger, at 2,997,000 calculated cosine similarities per language, per system, and within vs. across PoS boundaries.

\subsection{Pre-trained embeddings with noisy training data}\label{sec:updatingembeds}

When training a tagger with noisy training data and pre-trained embeddings, the question arises whether it is more beneficial to freeze the word embeddings or update them.
We hypothesize that freezing embeddings is  more beneficial in noisy training cases, as it helps to stabilize the signal from the pre-trained word embeddings while avoiding updates from the noisy training data. To test this hypothesis, we train the base tagger on high-quality gold training data (effectively, the UD training data sets), with and without freezing the word embeddings layer. We find that updating the word embedding layer is in fact beneficial in the high-quality training data regime: on average +0.4\% absolute improvement is obtained (mean over 21 languages). This is in sharp contrast to  the noisy training data regime, in which the baseline accuracy drops by as much as 1.2\% accuracy. Therefore, in this paper we train the tagger with pre-trained embeddings on projected WTC data and freeze the word embeddings lookup layer during training.


\section{Related work} 
\label{sec:related_work}

In recent years, natural language processing has witnessed a move towards deep learning approaches, in which automatic representation learning has become the de facto standard methodology~\cite{collobert2011natural,manning2015computational}. 

One of the first works that combines neural representations with semantic symbolic lexicons is the work on \textit{retrofitting}~\cite{retrofitting}. The main idea is to use the relations defined in semantic lexicons to refine word embedding representations, such that words linked in the lexical resource are encouraged to be closer to each other in the distributional space.

The majority of recent work on neural sequence prediction follows the commonly perceived wisdom that hand-crafted features are obsolete for deep learning methods. They rely on end-to-end training without resorting to additional linguistic resources. Our study contributes to the increasing literature to show the utility of linguistic resources for deep learning models. Most prior work in this direction can be found on machine translation~\cite{sennrich-haddow:2016:WMT,chen:ea:2017,Li:ea:2017,passban:ea:2018}, work on named entity recognition~\cite{wu:liu:cohn} and PoS tagging~\cite{sagot-martinezalonso:2017:IWPT}  who use lexicons, but as $n$-hot features and without examining the cross-lingual aspect. 

Somewhat complementary to evaluating  the utility of linguistic resources empirically is the increasing body of work that uses linguistic insights to try to understand what properties neural-based representations capture~\cite{kadar:ea:2017,adi:ea:2017,belinkov:ea:2017:acl,conneau:ea:2018,hupkes2018visualisation}. Shi et al. (2016) and Adi et al. (2017) introduced the idea of probing tasks (or `diagnostic classifiers'). Adi et al. (2017) evaluate several kinds of sentence encoders and propose a range of probing tasks around isolated aspects of sentence structure at the surface level (sentence length, word content and word order). This work has been  greatly expanded by including both syntactic and semantic probing tasks, careful sampling of probing task training data, and extending the framework to make it encoder agnostic~\cite{conneau:ea:2018}. A general observation here is that task-specific knowledge it needed in order to design relevant diagnostic tasks, which is not always straightforward.
For example, Gulordava~\shortcite{gulordava:ea:2018} investigate whether RNNs trained using a language model objective capture hierarchical syntactic information. They create nonsensical construction so that the RNN cannot rely on lexical or semantic clues, showing that RNNs still capture syntactic properties in sentence embeddings across the four tested languages while obfuscating lexical information. There is also more theoretical work on investigating the capabilities of recurrent neural networks, e.g., Weiss et al.~\shortcite{weiss} show that specific types of RNNs (LSTMs) are able to use counting mechanisms to recognize specific formal languages.

Finally, linguistic resources can also serve as proxy for evaluation. As recently shown~\cite{agic:ea:2017}, type-level information from dictionaries approximates PoS tagging accuracy in the absence of gold data for cross-lingual tagger evaluation. Their use of high-frequency word types inspired parts of our analysis.

\section{Conclusions}

We introduce and thoroughly analyze \textsc{DsDs}, a low-resource tagger that symbiotically leverages neural representations and symbolic linguistic knowledge by integrating them in a soft manner. 
The advantage of our approach is that embedding lexicons results in a tagger which is more resilient to noise, it can learn from limited lexicons and under missing distant sources. The more implicit use of the dictionary turns out to be more beneficial than approaches that rely more explicitly on symbolic knowledge, such a type constraints or retrofitting. \textsc{DsDs} can effectively learn from limited projected training data (5k instances), and freezing the pre-trained embeddings complement the learning signal well in this noisy data regime.
By analyzing the reliance of \textsc{DsDs} on the linguistic knowledge, we found that the composition of the lexicon is more important than its size. Our quantitative analysis also sheds light on the internal representations, showing that they get more sensitive to the task at hand, both via probing and the similarity analysis. This supports the empirical finding that the tagger learns from the lexicon beyond its coverage, and a ``take what you get'' approach is a viable strategy for low-resource tagging.

\bibliographystyle{authordate1}
\bibliography{bibliography}

\begin{thebibliography}{}

\bibitem[\protect\citename{Adi {\em et~al.\ }\relax, }2017]{adi:ea:2017}
Adi, Yossi, Kermany, Einat, Belinkov, Yonatan, Lavi, Ofer, \& Goldberg, Yoav.
  2017.
\newblock Fine-grained analysis of sentence embeddings using auxiliary
  prediction tasks.
\newblock {\em In:} {\em ICLR}.

\bibitem[\protect\citename{Agi\'{c} {\em et~al.\ }\relax, }2015]{agic:ea:2015}
Agi\'{c}, \v{Z}eljko, Hovy, Dirk, \& S{\o}gaard, Anders. 2015.
\newblock If all you have is a bit of the Bible: Learning POS taggers for truly
  low-resource languages.
\newblock {\em Pages  268--272 of:} {\em Proceedings of the 53rd Annual Meeting
  of the Association for Computational Linguistics and the 7th International
  Joint Conference on Natural Language Processing (Volume 2: Short Papers)}.
\newblock Beijing, China: Association for Computational Linguistics.

\bibitem[\protect\citename{Agi{\'c} {\em et~al.\ }\relax,
  }2016]{agic2016multilingual}
Agi{\'c}, {\v{Z}}eljko, Johannsen, Anders, Plank, Barbara, Mart{\'\i}nez,
  H{\'e}ctor~Alonso, Schluter, Natalie, \& S{\o}gaard, Anders. 2016.
\newblock Multilingual projection for parsing truly low-resource languages.
\newblock {\em Transactions of the Association for Computational Linguistics},
  {\bf 4}, 301--312.

\bibitem[\protect\citename{Agi{\'{c}} {\em et~al.\ }\relax,
  }2017]{agic:ea:2017}
Agi{\'{c}}, {\v{Z}}eljko, Plank, Barbara, \& S{\o}gaard, Anders. 2017.
\newblock Cross-lingual tagger evaluation without test data.
\newblock {\em Pages  248--253 of:} {\em Proceedings of the 15th Conference of
  the European Chapter of the Association for Computational Linguistics: Volume
  2, Short Papers}.
\newblock Association for Computational Linguistics.

\bibitem[\protect\citename{Al-Rfou {\em et~al.\ }\relax, }2013]{polyglot}
Al-Rfou, Rami, Perozzi, Bryan, \& Skiena, Steven. 2013.
\newblock Polyglot: Distributed word representations for multilingual nlp.
\newblock {\em arXiv preprint arXiv:1307.1662}.

\bibitem[\protect\citename{Belinkov {\em et~al.\ }\relax,
  }2017]{belinkov:ea:2017:acl}
Belinkov, Yonatan, Durrani, Nadir, Dalvi, Fahim, Sajjad, Hassan, \& Glass,
  James. 2017.
\newblock What do Neural Machine Translation Models Learn about Morphology?
\newblock {\em Pages  861--872 of:} {\em Proceedings of the 55th Annual Meeting
  of the Association for Computational Linguistics (Volume 1: Long Papers)}.
\newblock Association for Computational Linguistics.

\bibitem[\protect\citename{Björnsson, }1983]{bjornsson1983readability}
Björnsson, C.~H. 1983.
\newblock Readability of Newspapers in 11 Languages.
\newblock {\em Reading Research Quarterly}, {\bf 18}(4), 480--497.

\bibitem[\protect\citename{Bohnet {\em et~al.\ }\relax,
  }2018]{bohnet2018morphosyntactic}
Bohnet, Bernd, McDonald, Ryan, Simoes, Goncalo, Andor, Daniel, Pitler, Emily,
  \& Maynez, Joshua. 2018.
\newblock Morphosyntactic Tagging with a Meta-BiLSTM Model over Context
  Sensitive Token Encodings.
\newblock {\em arXiv preprint arXiv:1805.08237}.

\bibitem[\protect\citename{Bojanowski {\em et~al.\ }\relax,
  }2016]{bojanowski2016enriching}
Bojanowski, Piotr, Grave, Edouard, Joulin, Armand, \& Mikolov, Tomas. 2016.
\newblock Enriching Word Vectors with Subword Information.
\newblock {\em arXiv preprint arXiv:1607.04606}.

\bibitem[\protect\citename{Buchholz \& Marsi,
  }2006]{buchholz-marsi:2006:CoNLL-X}
Buchholz, Sabine, \& Marsi, Erwin. 2006.
\newblock CoNLL-X Shared Task on Multilingual Dependency Parsing.
\newblock {\em Pages  149--164 of:} {\em Proceedings of the Tenth Conference on
  Computational Natural Language Learning (CoNLL-X)}.
\newblock New York City: Association for Computational Linguistics.

\bibitem[\protect\citename{Chen {\em et~al.\ }\relax, }2017]{chen:ea:2017}
Chen, Huadong, Huang, Shujian, Chiang, David, \& Chen, Jiajun. 2017.
\newblock Improved Neural Machine Translation with a Syntax-Aware Encoder and
  Decoder.
\newblock {\em Pages  1936--1945 of:} {\em Proceedings of the 55th Annual
  Meeting of the Association for Computational Linguistics (Volume 1: Long
  Papers)}.
\newblock Association for Computational Linguistics.

\bibitem[\protect\citename{Collobert {\em et~al.\ }\relax,
  }2011]{collobert2011natural}
Collobert, Ronan, Weston, Jason, Bottou, L{\'e}on, Karlen, Michael,
  Kavukcuoglu, Koray, \& Kuksa, Pavel. 2011.
\newblock Natural language processing (almost) from scratch.
\newblock {\em Journal of Machine Learning Research}, {\bf 12}(Aug),
  2493--2537.

\bibitem[\protect\citename{Conneau {\em et~al.\ }\relax,
  }2018]{conneau:ea:2018}
Conneau, Alexis, Kruszewski, Germ{\'a}n, Lample, Guillaume, Barrault, Lo{\"i}c,
  \& Baroni, Marco. 2018.
\newblock What you can cram into a single .. vector: Probing sentence
  embeddings for linguistic properties.
\newblock {\em Pages  2126--2136 of:} {\em Proceedings of the 56th Annual
  Meeting of the Association for Computational Linguistics (Volume 1: Long
  Papers)}.
\newblock Association for Computational Linguistics.

\bibitem[\protect\citename{Das \& Petrov, }2011]{das-petrov:2011}
Das, Dipanjan, \& Petrov, Slav. 2011.
\newblock Unsupervised Part-of-Speech Tagging with Bilingual Graph-Based
  Projections.
\newblock {\em Pages  600--609 of:} {\em Proceedings of the 49th Annual Meeting
  of the Association for Computational Linguistics: Human Language
  Technologies}.
\newblock Portland, Oregon, USA: Association for Computational Linguistics.

\bibitem[\protect\citename{Enghoff {\em et~al.\ }\relax, }2018]{enghoff2018low}
Enghoff, Jan~Vium, Harrison, S{\o}ren, \& Agi{\'{c}}, {\v{Z}}eljko. 2018.
\newblock Low-resource named entity recognition via multi-source projection:
  Not quite there yet?
\newblock {\em Pages  195--201 of:} {\em Proceedings of the 2018 EMNLP Workshop
  W-NUT: The 4th Workshop on Noisy User-generated Text}.
\newblock Association for Computational Linguistics.

\bibitem[\protect\citename{Evang \& Bos, }2016]{evang2016cross}
Evang, Kilian, \& Bos, Johan. 2016.
\newblock Cross-lingual learning of an open-domain semantic parser.
\newblock {\em Pages  579--588 of:} {\em Proceedings of COLING 2016, the 26th
  International Conference on Computational Linguistics: Technical Papers}.

\bibitem[\protect\citename{Faruqui {\em et~al.\ }\relax, }2015]{retrofitting}
Faruqui, Manaal, Dodge, Jesse, Jauhar, Sujay~Kumar, Dyer, Chris, Hovy, Eduard,
  \& Smith, Noah~A. 2015.
\newblock Retrofitting Word Vectors to Semantic Lexicons.
\newblock {\em Pages  1606--1615 of:} {\em Proceedings of the 2015 Conference
  of the North American Chapter of the Association for Computational
  Linguistics: Human Language Technologies}.
\newblock Association for Computational Linguistics.

\bibitem[\protect\citename{Garrette \& Baldridge,
  }2013]{garrette-baldridge:2013:NAACL-HLT}
Garrette, Dan, \& Baldridge, Jason. 2013.
\newblock Learning a Part-of-Speech Tagger from Two Hours of Annotation.
\newblock {\em Pages  138--147 of:} {\em Proceedings of the 2013 Conference of
  the North American Chapter of the Association for Computational Linguistics:
  Human Language Technologies}.
\newblock Atlanta, Georgia: Association for Computational Linguistics.

\bibitem[\protect\citename{Graves \& Schmidhuber,
  }2005]{graves:schmidhuber:2005}
Graves, Alex, \& Schmidhuber, J{\"u}rgen. 2005.
\newblock Framewise phoneme classification with bidirectional LSTM and other
  neural network architectures.
\newblock {\em Neural Networks}, {\bf 18}(5), 602--610.

\bibitem[\protect\citename{Gulordava {\em et~al.\ }\relax,
  }2018]{gulordava:ea:2018}
Gulordava, Kristina, Bojanowski, Piotr, Grave, Edouard, Linzen, Tal, \& Baroni,
  Marco. 2018.
\newblock Colorless Green Recurrent Networks Dream Hierarchically.
\newblock {\em Pages  1195--1205 of:} {\em Proceedings of the 2018 Conference
  of the North American Chapter of the Association for Computational
  Linguistics: Human Language Technologies, Volume 1 (Long Papers)}.
\newblock Association for Computational Linguistics.

\bibitem[\protect\citename{Hochreiter \& Schmidhuber,
  }1997]{Hochreiter:Schmidhuber:97}
Hochreiter, Sepp, \& Schmidhuber, J{\"u}rgen. 1997.
\newblock Long short-term memory.
\newblock {\em Neural computation}, {\bf 9}(8), 1735--1780.

\bibitem[\protect\citename{Hupkes {\em et~al.\ }\relax,
  }2018]{hupkes2018visualisation}
Hupkes, Dieuwke, Veldhoen, Sara, \& Zuidema, Willem. 2018.
\newblock Visualisation and'diagnostic classifiers' reveal how recurrent and
  recursive neural networks process hierarchical structure.
\newblock {\em Journal of Artificial Intelligence Research}, {\bf 61},
  907--926.

\bibitem[\protect\citename{K{\'a}d{\'a}r {\em et~al.\ }\relax,
  }2017]{kadar:ea:2017}
K{\'a}d{\'a}r, {\'A}kos, Chrupa{\l}a, Grzegorz, \& Alishahi, Afra. 2017.
\newblock Representation of Linguistic Form and Function in Recurrent Neural
  Networks.
\newblock {\em Computational Linguistics}, {\bf 43}(4), 761--780.

\bibitem[\protect\citename{Kiperwasser \& Goldberg, }2016]{kiperwasser2016}
Kiperwasser, Eliyahu, \& Goldberg, Yoav. 2016.
\newblock Simple and accurate dependency parsing using bidirectional LSTM
  feature representations.
\newblock {\em arXiv preprint arXiv:1603.04351}.

\bibitem[\protect\citename{Kirov {\em et~al.\ }\relax, }2016]{KIROV16.1077}
Kirov, Christo, Sylak-Glassman, John, Que, Roger, \& Yarowsky, David. 2016.
\newblock Very-large Scale Parsing and Normalization of Wiktionary
  Morphological Paradigms.
\newblock {\em In:} Chair), Nicoletta Calzolari~(Conference, Choukri, Khalid,
  Declerck, Thierry, Goggi, Sara, Grobelnik, Marko, Maegaard, Bente, Mariani,
  Joseph, Mazo, Helene, Moreno, Asuncion, Odijk, Jan, \& Piperidis, Stelios
  (eds), {\em Proceedings of the Tenth International Conference on Language
  Resources and Evaluation (LREC 2016)}.
\newblock Paris, France: European Language Resources Association (ELRA).

\bibitem[\protect\citename{Li {\em et~al.\ }\relax, }2017]{Li:ea:2017}
Li, Junhui, Xiong, Deyi, Tu, Zhaopeng, Zhu, Muhua, Zhang, Min, \& Zhou,
  Guodong. 2017.
\newblock Modeling Source Syntax for Neural Machine Translation.
\newblock {\em Pages  688--697 of:} {\em Proceedings of the 55th Annual Meeting
  of the Association for Computational Linguistics (Volume 1: Long Papers)}.
\newblock Vancouver, Canada: Association for Computational Linguistics.

\bibitem[\protect\citename{Li {\em et~al.\ }\relax, }2012]{li-et-al:2012}
Li, Shen, Gra\c{c}a, Jo\~{a}o, \& Taskar, Ben. 2012.
\newblock Wiki-ly Supervised Part-of-Speech Tagging.
\newblock {\em Pages  1389--1398 of:} {\em Proceedings of the 2012 Joint
  Conference on Empirical Methods in Natural Language Processing and
  Computational Natural Language Learning}.
\newblock Jeju Island, Korea: Association for Computational Linguistics.

\bibitem[\protect\citename{Mann \& Whitney, }1947]{mann1947test}
Mann, Henry~B, \& Whitney, Donald~R. 1947.
\newblock On a test of whether one of two random variables is stochastically
  larger than the other.
\newblock {\em The annals of mathematical statistics},  50--60.

\bibitem[\protect\citename{Manning, }2015]{manning2015computational}
Manning, Christopher~D. 2015.
\newblock Computational linguistics and deep learning.
\newblock {\em Computational Linguistics}, {\bf 41}(4), 701--707.

\bibitem[\protect\citename{Mayhew {\em et~al.\ }\relax, }2017]{mayhew2017cheap}
Mayhew, Stephen, Tsai, Chen-Tse, \& Roth, Dan. 2017.
\newblock Cheap Translation for Cross-Lingual Named Entity Recognition.
\newblock {\em Pages  2536--2545 of:} {\em Proceedings of the 2017 Conference
  on Empirical Methods in Natural Language Processing}.
\newblock Association for Computational Linguistics.

\bibitem[\protect\citename{Ni {\em et~al.\ }\relax, }2017]{ni2017weakly}
Ni, Jian, Dinu, Georgiana, \& Florian, Radu. 2017.
\newblock Weakly Supervised Cross-Lingual Named Entity Recognition via
  Effective Annotation and Representation Projection.
\newblock {\em Pages  1470--1480 of:} {\em Proceedings of the 55th Annual
  Meeting of the Association for Computational Linguistics (Volume 1: Long
  Papers)}.
\newblock Association for Computational Linguistics.

\bibitem[\protect\citename{Nivre {\em et~al.\ }\relax,
  }2007]{nivre-EtAl:2007:EMNLP-CoNLL2007}
Nivre, Joakim, Hall, Johan, K\"ubler, Sandra, McDonald, Ryan, Nilsson, Jens,
  Riedel, Sebastian, \& Yuret, Deniz. 2007.
\newblock The {CoNLL} 2007 Shared Task on Dependency Parsing.
\newblock {\em Pages  915--932 of:} {\em Proceedings of the CoNLL Shared Task
  Session of EMNLP-CoNLL 2007}.
\newblock Prague, Czech Republic: Association for Computational Linguistics.

\bibitem[\protect\citename{Nivre {\em et~al.\ }\relax, }2017]{ud21}
Nivre, Joakim, Agi{\'c}, {\v Z}eljko, \& Ahrenberg~et al., Lars. 2017.
\newblock {\em Universal Dependencies 2.1}.
\newblock {LINDAT}/{CLARIN} digital library at the Institute of Formal and
  Applied Linguistics ({{\'U}FAL}), Faculty of Mathematics and Physics, Charles
  University.

\bibitem[\protect\citename{Passban {\em et~al.\ }\relax,
  }2018]{passban:ea:2018}
Passban, Peyman, Liu, Qun, \& Way, Andy. 2018.
\newblock Improving Character-Based Decoding Using Target-Side Morphological
  Information for Neural Machine Translation.
\newblock {\em Pages  58--68 of:} {\em Proceedings of the 2018 Conference of
  the North American Chapter of the Association for Computational Linguistics:
  Human Language Technologies, Volume 1 (Long Papers)}.
\newblock Association for Computational Linguistics.

\bibitem[\protect\citename{Petrov {\em et~al.\ }\relax, }2012]{petrov:ea:2012}
Petrov, Slav, Das, Dipanjan, \& McDonald, Ryan. 2012.
\newblock A Universal Part-of-Speech Tagset.
\newblock {\em In:} Chair), Nicoletta Calzolari~(Conference, Choukri, Khalid,
  Declerck, Thierry, Doğan, Mehmet~Uğur, Maegaard, Bente, Mariani, Joseph,
  Moreno, Asuncion, Odijk, Jan, \& Piperidis, Stelios (eds), {\em Proceedings
  of the Eight International Conference on Language Resources and Evaluation
  (LREC'12)}.
\newblock Istanbul, Turkey: European Language Resources Association (ELRA).

\bibitem[\protect\citename{Plank \& Agi{\'{c}}, }2018]{dsds}
Plank, Barbara, \& Agi{\'{c}}, {\v{Z}}eljko. 2018.
\newblock Distant Supervision from Disparate Sources for Low-Resource
  Part-of-Speech Tagging.
\newblock {\em Pages  614--620 of:} {\em Proceedings of the 2018 Conference on
  Empirical Methods in Natural Language Processing}.
\newblock Association for Computational Linguistics.

\bibitem[\protect\citename{Plank {\em et~al.\ }\relax, }2016]{plank:ea:2016}
Plank, Barbara, S{\o}gaard, Anders, \& Goldberg, Yoav. 2016.
\newblock Multilingual Part-of-Speech Tagging with Bidirectional Long
  Short-Term Memory Models and Auxiliary Loss.
\newblock {\em Pages  412--418 of:} {\em Proceedings of the 54th Annual Meeting
  of the Association for Computational Linguistics (Volume 2: Short Papers)}.
\newblock Association for Computational Linguistics.

\bibitem[\protect\citename{Rasooli \& Collins, }2016]{rasooli2016cross}
Rasooli, Mohammad~Sadegh, \& Collins, Michael. 2016.
\newblock Cross-lingual syntactic transfer with limited resources.
\newblock {\em arXiv preprint arXiv:1610.06227}.

\bibitem[\protect\citename{Ruder {\em et~al.\ }\relax, }2017]{ruder2017survey}
Ruder, Sebastian, Vuli{\'c}, Ivan, \& S{\o}gaard, Anders. 2017.
\newblock A survey of cross-lingual word embedding models.
\newblock {\em arXiv preprint arXiv:1706.04902}.

\bibitem[\protect\citename{Sagot \& Mart\'{i}nez~Alonso,
  }2017]{sagot-martinezalonso:2017:IWPT}
Sagot, Beno\^{i}t, \& Mart\'{i}nez~Alonso, H\'{e}ctor. 2017.
\newblock Improving neural tagging with lexical information.
\newblock {\em Pages  25--31 of:} {\em Proceedings of the 15th International
  Conference on Parsing Technologies}.
\newblock Pisa, Italy: Association for Computational Linguistics.

\bibitem[\protect\citename{Sennrich \& Haddow, }2016]{sennrich-haddow:2016:WMT}
Sennrich, Rico, \& Haddow, Barry. 2016.
\newblock Linguistic Input Features Improve Neural Machine Translation.
\newblock {\em Pages  83--91 of:} {\em Proceedings of the First Conference on
  Machine Translation}.
\newblock Berlin, Germany: Association for Computational Linguistics.

\bibitem[\protect\citename{Shi {\em et~al.\ }\relax, }2016]{shi:ea:2016}
Shi, Xing, Padhi, Inkit, \& Knight, Kevin. 2016.
\newblock Does String-Based Neural MT Learn Source Syntax?
\newblock {\em Pages  1526--1534 of:} {\em Proceedings of the 2016 Conference
  on Empirical Methods in Natural Language Processing}.
\newblock Association for Computational Linguistics.

\bibitem[\protect\citename{Sperlich {\em et~al.\ }\relax,
  }2015]{sperlich2015perceptual}
Sperlich, Anja, Schad, Daniel~J., \& Laubrock, Jochen. 2015.
\newblock When preview information starts to matter: Development of the
  perceptual span in German beginning readers.
\newblock {\em Journal of Cognitive Psychology}, {\bf 27}(5), 511--530.

\bibitem[\protect\citename{Stratos {\em et~al.\ }\relax,
  }2016]{stratos2016unsupervised}
Stratos, Karl, Collins, Michael, \& Hsu, Daniel. 2016.
\newblock Unsupervised Part-Of-Speech Tagging with Anchor Hidden Markov Models.
\newblock {\em Transactions of the Association for Computational Linguistics},
  {\bf 4}, 245--257.

\bibitem[\protect\citename{T{\"a}ckstr{\"o}m {\em et~al.\ }\relax,
  }2013]{tackstrom:ea:2013}
T{\"a}ckstr{\"o}m, Oscar, Das, Dipanjan, Petrov, Slav, McDonald, Ryan, \&
  Nivre, Joakim. 2013.
\newblock Token and type constraints for cross-lingual part-of-speech tagging.
\newblock {\em Transactions of the Association for Computational Linguistics},
  {\bf 1}, 1--12.

\bibitem[\protect\citename{Tiedemann \& Agi{\'c},
  }2016]{tiedemann2016synthetic}
Tiedemann, J{\"o}rg, \& Agi{\'c}, Zeljko. 2016.
\newblock Synthetic treebanking for cross-lingual dependency parsing.
\newblock {\em Journal of Artificial Intelligence Research}, {\bf 55},
  209--248.

\bibitem[\protect\citename{Weiss {\em et~al.\ }\relax, }2018]{weiss}
Weiss, Gail, Goldberg, Yoav, \& Yahav, Eran. 2018.
\newblock On the Practical Computational Power of Finite Precision RNNs for
  Language Recognition.
\newblock {\em Pages  740--745 of:} {\em Proceedings of the 56th Annual Meeting
  of the Association for Computational Linguistics (Volume 2: Short Papers)}.
\newblock Association for Computational Linguistics.

\bibitem[\protect\citename{Wu {\em et~al.\ }\relax, }2018]{wu:liu:cohn}
Wu, Minghao, Liu, Fei, \& Cohn, Trevor. 2018.
\newblock Evaluating the Utility of Hand-crafted Features in Sequence
  Labelling.
\newblock {\em Pages  2850--2856 of:} {\em Proceedings of the 2018 Conference
  on Empirical Methods in Natural Language Processing}.
\newblock Association for Computational Linguistics.

\bibitem[\protect\citename{Yang {\em et~al.\ }\relax, }2018]{yang:2018:coling}
Yang, Jie, Liang, Shuailong, \& Zhang, Yue. 2018.
\newblock Design Challenges and Misconceptions in Neural Sequence Labeling.
\newblock {\em Pages  3879--3889 of:} {\em Proceedings of the 27th
  International Conference on Computational Linguistics}.
\newblock Association for Computational Linguistics.

\bibitem[\protect\citename{Yarowsky {\em et~al.\ }\relax,
  }2001]{yarowsky2001inducing}
Yarowsky, David, Ngai, Grace, \& Wicentowski, Richard. 2001.
\newblock Inducing multilingual text analysis tools via robust projection
  across aligned corpora.
\newblock {\em Pages  1--8 of:} {\em Proceedings of the first international
  conference on Human language technology research}.
\newblock Association for Computational Linguistics.

\bibitem[\protect\citename{Yu {\em et~al.\ }\relax, }2016]{zeman2016bible}
Yu, Zhiwei, Mareček, David, Žabokrtský, Zdeněk, \& Zeman, Daniel. 2016.
\newblock If You Even Don't Have a Bit of Bible: Learning Delexicalized POS
  Taggers.
\newblock {\em In:} Chair), Nicoletta Calzolari~(Conference, Choukri, Khalid,
  Declerck, Thierry, Goggi, Sara, Grobelnik, Marko, Maegaard, Bente, Mariani,
  Joseph, Mazo, Helene, Moreno, Asuncion, Odijk, Jan, \& Piperidis, Stelios
  (eds), {\em Proceedings of the Tenth International Conference on Language
  Resources and Evaluation (LREC 2016)}.
\newblock Paris, France: European Language Resources Association (ELRA).

\bibitem[\protect\citename{Zeman {\em et~al.\ }\relax, }2014]{zeman2014hamledt}
Zeman, Daniel, Du{\v{s}}ek, Ond{\v{r}}ej, Mare{\v{c}}ek, David, Popel, Martin,
  Ramasamy, Loganathan, {\v{S}}t{\v{e}}p{\'a}nek, Jan, {\v{Z}}abokrtsk{\`y},
  Zden{\v{e}}k, \& Haji{\v{c}}, Jan. 2014.
\newblock HamleDT: Harmonized multi-language dependency treebank.
\newblock {\em Language Resources and Evaluation}, {\bf 48}(4), 601--637.

\end{thebibliography}

\label{lastpage}

\end{document}